\documentclass{article}

\usepackage{microtype}
\usepackage{graphicx}
\usepackage{subcaption}
\usepackage{booktabs} 
\usepackage{amsmath}
\usepackage{pifont}
\usepackage{lipsum}
\usepackage{xcolor}

\usepackage{hyperref}
\usepackage{titlesec}
\usepackage{caption}

\usepackage[ruled,lined]{algorithm2e}

\SetCommentSty{mycommfont}

\newcommand{\red}[1]{\textcolor{red}{#1}}
\newcommand{\blue}[1]{\textcolor{blue}{#1}}

\newcommand{\arxcg}[1]{#1}

\usepackage[T1]{fontenc}
\usepackage{array}
\usepackage[accepted]{icml2024}

\usepackage{amsmath}
\usepackage{amssymb}
\usepackage{mathtools}
\usepackage{amsthm}
\usepackage{graphicx}

\usepackage[capitalize,noabbrev]{cleveref}

\theoremstyle{plain}

\theoremstyle{definition}

\theoremstyle{remark}

\usepackage[textsize=tiny]{todonotes}
\usepackage{caption}

\DeclareMathOperator*{\argmin}{arg\,min}  

\icmltitlerunning{Dr.~Strategy: Model-Based Generalist Agents with Strategic Dreaming}

\begin{document}

\twocolumn[
\icmltitle{Dr.~Strategy: Model-Based Generalist Agents with Strategic Dreaming}



\icmlsetsymbol{equal}{*}

\begin{icmlauthorlist}
\icmlauthor{Hany Hamed}{equal,kaist}
\icmlauthor{Subin Kim}{equal,kaist}
\icmlauthor{Dongyeong Kim}{kaist}
\icmlauthor{Jaesik Yoon}{kaist,sap}
\icmlauthor{Sungjin Ahn}{kaist}
\end{icmlauthorlist}

\icmlaffiliation{kaist}{KAIST}
\icmlaffiliation{sap}{SAP}

\icmlcorrespondingauthor{Sungjin Ahn}{sjn.ahn@gmail.com}

\icmlkeywords{Model-Based Reinforcement Learning, Image-Based Reinforcement Learning, Goal-Conditioned Reinforcement Learning, Machine Learning, ICML}

\vskip 0.3in
]



\printAffiliationsAndNotice{\icmlEqualContribution} 

\begin{abstract}
Model-based reinforcement learning (MBRL) has been a primary approach to ameliorating the sample efficiency issue as well as to make a generalist agent. However, there has not been much effort toward enhancing the strategy of dreaming itself. Therefore, it is a question \textit{whether and how an agent can ``\textit{dream better}''} in a more structured and strategic way. In this paper, inspired by the observation from cognitive science suggesting that humans use a spatial divide-and-conquer strategy in planning, we propose a new MBRL agent, called \textbf{Dr.~Strategy}, which is equipped with a novel \textbf{Dr}eaming \textbf{Strategy}. The proposed agent realizes a version of divide-and-conquer-like strategy in dreaming. This is achieved by learning a set of latent landmarks and then utilizing these to learn a landmark-conditioned highway policy. With the highway policy, the agent can first learn in the dream to move to a landmark, and from there it tackles the exploration and achievement task in a more focused way. In experiments, we show that the proposed model outperforms prior pixel-based MBRL methods in various visually complex and partially observable navigation tasks. 

\end{abstract}
\section{Introduction}

A crucial capability of generalist agents, such as humans, is to explore environments and acquire the skills needed to achieve various goals, continuously and in an open-ended way. It is particularly important for these agents to become efficient \arxcg{explorers} and achievers in an unsupervised or self-supervised manner. It enables them to survive and become more competent in a more scalable way as well as in more flexible open-ended environments, where future tasks aren't predefined but can evolve over time. 

This capability is equally important for artificial generalist agents, such as Reinforcement Learning (RL) agents \cite{sutton_rl}, including robots and virtual agents in games like Minecraft \cite{minerl}. However, these artificial agents currently have a significant limitation compared to humans: low sample efficiency. They require much more experience data than humans \cite{dqn,a3c}. Considering these agents could operate in a real-time physical world and are susceptible to physical damage, improving sample efficiency is of top priority. It is particularly more challenging in more realistic settings where observations are high-dimensional (e.g., images) and partially observable \cite{openai_five,dm_starcraft2}.

\begin{figure}
  \centering
  \includegraphics[width=1\linewidth]{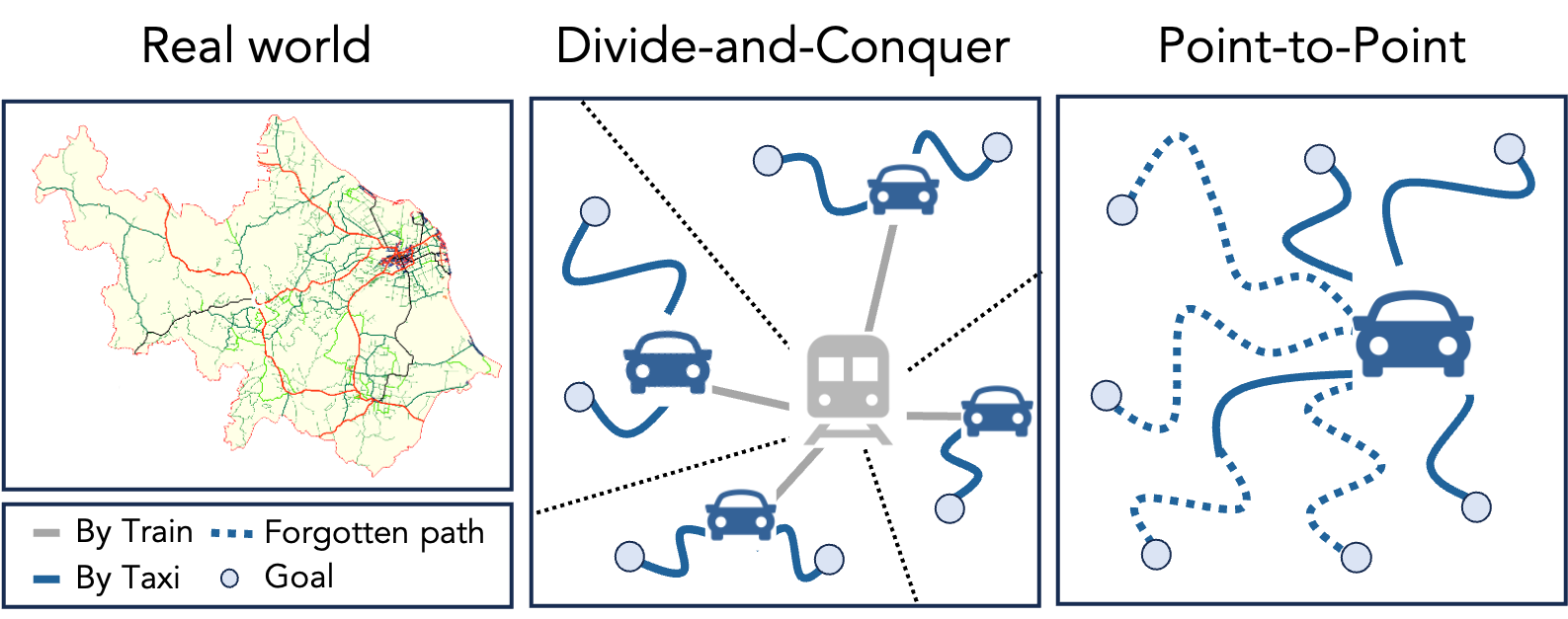}
  \vspace{-1.5em}
  
  \caption{\textbf{(Left)} In the real world, humans maintain a hierarchical spatial structure for easy navigation. \textbf{(Right)} Trying to memorize all the streets on the map can lead to an overwhelming amount of information, making it difficult to retain the information effectively. \textbf{(Middle)} In contrast, choosing to travel by train to move between cities and transfer to a taxi at the terminal minimizes the complexity, allowing one to concentrate on local routes starting from the terminal near the destination.}
  \vspace{-1.5em}
  \label{fig:example}
\end{figure}
Currently, a primary approach in RL to improving sample efficiency is via model-based reinforcement learning (MBRL) \cite{sutton1991dyna,worldmodel, hafner2020dreamerv2}. In this approach, the agent uses experience data to learn both the representation of the observations and states as well as the transition dynamics of the environment, known as a world model. This enables the agent to learn its policy within an internal model of the world instead of the real world via planning (or, simulation or dreaming).
An example of such an unsupervised model-based generalist agent is LEXA \cite{mendonca2021lexa}. 

On the other hand, research in cognitive science suggests that humans use structured and strategic planning, such as spatial divide-and-conquer, when tackling complex problems \cite{chun1998contextual}. For example, when navigating to a specific location, humans typically break down the task into two stages: first, they plan to reach a familiar landmark near the destination, then they use a local and focused strategy to get from that landmark to the target, as shown in Figure~\ref{fig:example}.~This \textit{divide-and-conquer}-like approach is effective as it reduces the space to learn. Without this, it would require to learn all point-to-point navigation paths separately, requiring a lot of experience data. However, in current MBRL agents like LEXA, the process of dreaming or imagination is guided by a rather naive strategy such as random i.i.d.~sampling from the replay buffer. 

In this paper, we raise the following questions: ``Is more structured and strategic dreaming possible?'', if so, ``how could we implement this idea in the modern MBRL frameworks?” and ``how could this improve generalist agents?'' 
To this end, we propose a strategic model-based generalist agent, \textit{Dr.~Strategy} (short for ``\textbf{\textit{Dr}}eam \textbf{\textit{Strategy}}”). 
Our key idea is that a divide-and-conquer approach leveraging the structure of \textit{latent landmarks} can enhance the efficiency of dreaming in MBRL and promote better exploration and achievement quality of a generalist agent.

The proposed model consists of four main modules. First, to obtain landmarks, we map each state from the replay buffer to a discrete representation called \textit{landmarks} through VQ-VAE \cite{razavi2019generating_vqvae2}. Second, we train a landmark-conditioned policy called \textit{highway policy},  specialized to move only to landmarks instead of arbitrary position, unlike goal-conditioned policy. Thirdly, we train an exploration policy (Explorer) and a goal-conditioned policy (Achiever) through dreaming. However, unlike LEXA, the two policies take advantage of starting from beneficial landmarks selected from \textit{strategic dreaming and planning}. 
Thus, they solve the problem locally in a focused way, following the highway policy to bring the agent to the selected landmark. This realizes the divide-and-conquer-like approach. 
In experiments, we show that the proposed model outperforms prior pixel-based MBRL methods in various visually complex and partially observable navigation tasks, while also showing comparative results in robot manipulation tasks.

The main contributions of this paper are as follows. We propose the concept of ``strategic dreaming" in pixel-based MBRL in the sense that the agent can leverage the structure of the state space such as landmarks to enable a divide-and-conquer-like strategy during dreaming, and then propose the first MBRL agent to realize and demonstrate the benefits of this concept. We also provide empirical evidence that this approach can enhance the accuracy and efficiency of MBRL agents in the generalist setting similar to LEXA. Additionally, we also introduce a set of benchmarks for visually complex navigation tasks.

\section{Dr.~Strategy Agent}
\begin{figure*}[!t]
\centering 
\begin{subfigure}{\linewidth}
   \centering
    \includegraphics[width=0.9\linewidth]{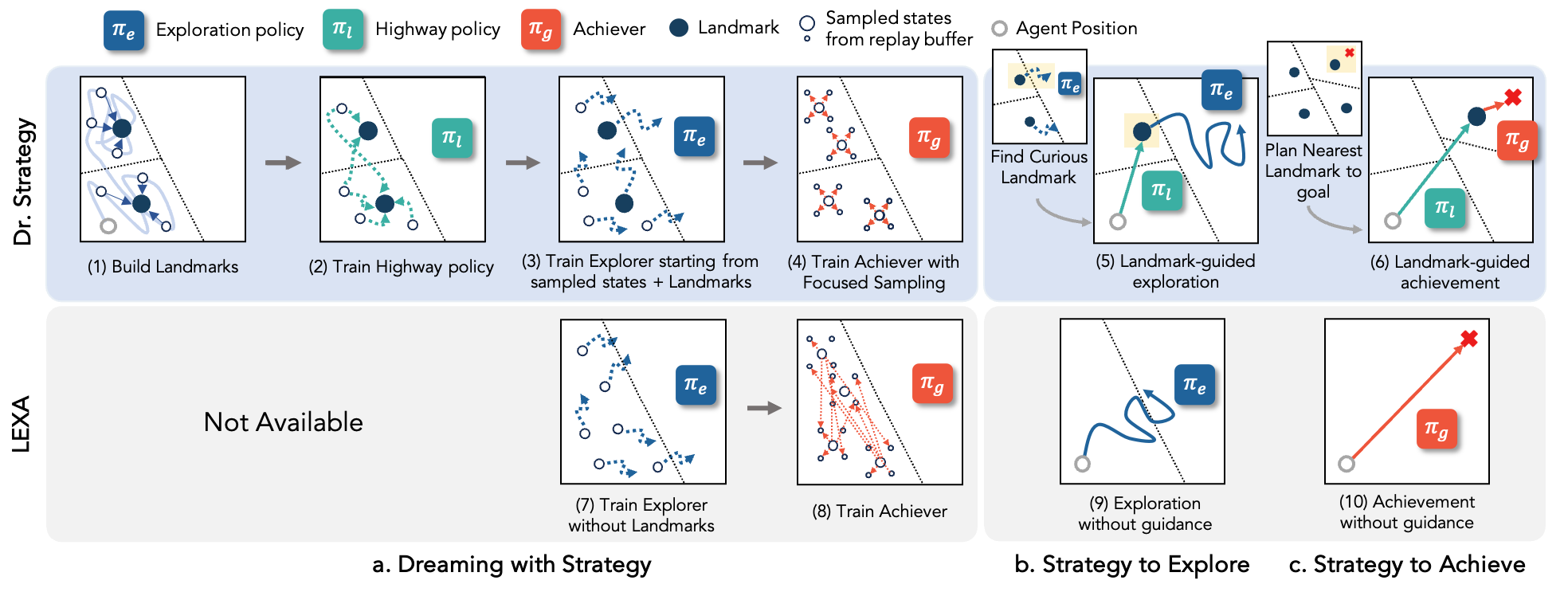}
    \vspace{-.7em}
    \label{fig:overview}
\end{subfigure}
\begin{subfigure}{\linewidth}
   \centering
   \includegraphics[width=0.9\linewidth]
   {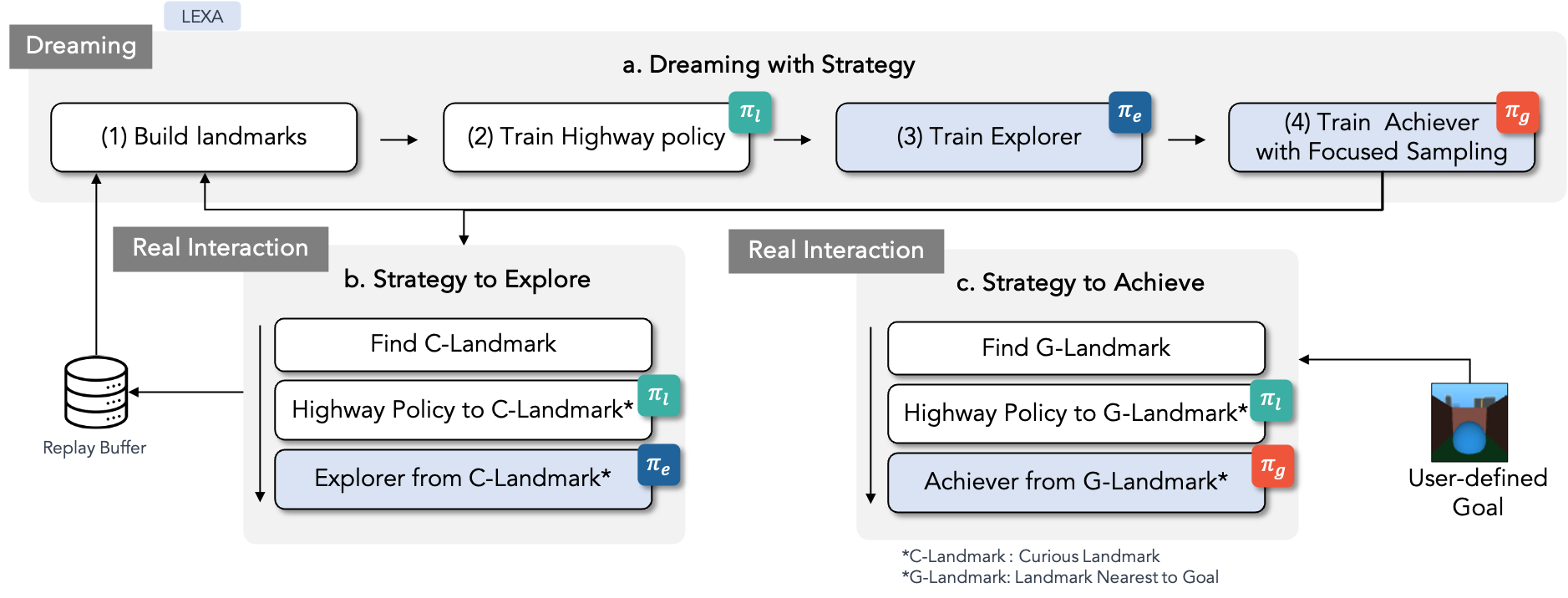}
   \vspace{-1.5em}
    \label{fig:schema_subfig}
\end{subfigure}
\caption{\textbf{Comparison between Dr.~Strategy and LEXA.} \textbf{a.} We construct latent landmarks and train Highway policy $\pi_{l}(a_t|s_t, l)$, Explorer $\pi_{e}(a_t|s_t)$, and Achiever $\pi_g(a_t|s_t, e_g)$ in imagination. The Achiever is trained by Focused Sampling, which is conditioning goals within a small number of steps instead of random sampling. All three policies are purely trained with imagined trajectories from the world model. \textbf{b.} During exploration, we only evaluate the landmarks, and call the landmark with the highest exploration potential ``Curious Landmark" (C-Landmark). In a real environment, the Highway policy moves to the curious landmark, and the Explorer resumes exploration. The agent iterates training and exploration with a certain frequency $T_F$. \textbf{c.} During test time, we find the landmark that is nearest to the given pixel-level goal (G-Landmark). The Highway policy reaches G-Landmark, and the Achiever proceeds to achieve the goal immediately after. The blue boxes in the bottom half of the figure indicate the modules of LEXA, which are Explorer and Achiever without focused sampling and landmarks.}
\vspace{-1.3em}
\label{fig:schema}
\end{figure*}
To enable a structured divide-and-conquer approach and thus enhance the efficiency of dreaming in world models for goal-conditioned agents, we introduce our proposed model, \textit{Dr.~Strategy}. A key change to prior model-based goal-conditioned approaches is the use of \textit{latent landmarks}. 
Latent landmarks are a set of latent states representing the experience of the agent, which enables the agent to strategically focus on essential information and thus dream structurally.
In our proposed model, we \textit{divide} our experience via landmarks and \textit{conquer} by starting from the landmarks, thereby guiding the agent to explore and achieve goals efficiently and with precision. We call the overall process of training and planning to exploit the divide-and-conquer strategy \textit{``Strategic Dreaming''}.

Dr.~Strategy consists of three policies: \arxcg{the \textit{Highway policy}, which helps reach landmarks; \textit{Explorer}, which explores distant points using the world model; and \textit{Achiever}, which reaches specified goals in divided areas.} Additionally, we incorporate \textit{Focused Sampling} during Achiever training to increase accuracy.
As illustrated in Figure~\ref{fig:schema}, our approach consists of two phases: \arxcg{(1) We construct latent landmarks from the explored states (Section~\ref{sec:landmarks}), train the three policies in imagination through \textit{Strategic Dreaming} (Section~\ref{sec:building_blocks}), and then explore through curious landmark-guided exploration (Section~\ref{sec:discovery}).} (2) We then achieve downstream tasks in the real environment exploiting the Highway policy and Achiever (Section~\ref{sec:conquer}).

\subsection{World Model}


To enhance the accurate prediction by high-dimensional pixel-level inputs, we employ a Recurrent State Space Model (RSSM) \cite{hafner2019rssm}. 
The world model works as a virtual simulator, predicting the transition dynamics \arxcg{of the real environment. The policy interacts with the imagined trajectories generated in parallel by sampling from the world model. We refer to this as \textit{``Dreaming"}. Thus, we can train policies using the imagined trajectories instead of interacting directly with the real environment (refer to Appendix \ref{appendix:baselines} for more details).}
The components comprising the world model include:
\begin{align}
\text{Dynamics}: & \quad \hat{s}_t = \texttt{dyn}_{\theta}(s_{t-1}, a_{t-1}) \\
\text{Representation}: & \quad s_t = \texttt{repr}_{\theta}(s_{t-1}, a_{t-1}, x_t) \\
\text{Encoder}: & \quad e_t = \texttt{enc}_{\theta}(x_t) \\
\text{Decoder}: & \quad \hat{x}_t = \texttt{dec}_{\theta}(s_t), 
\end{align}
where $s_t$ is the model state which is constructed as a concatenation of a deterministic state from GRU \cite{cho2014learning} and a discrete stochastic state \cite{hafner2020dreamerv2}. $a_t$ and $x_t$ are action and observation, respectively.
The world model is trained by optimizing the evidence lower bound (ELBO) through stochastic backpropagation \cite{kingma2013auto, rezende2014stochastic} using the Adam optimizer \cite{kingma2014adam}.

\subsection{Building Latent Landmarks}
\label{sec:landmarks}
We project the model states onto discrete $N$ codes in the codebook we call \textit{landmarks} using the VQ-VAE~\cite{van2017vqvae1}.
Landmarks can be seen as cluster centers partitioning the state space into a number of codes in the codebook.
To exploit these landmarks, we train the latent landmark-conditioned policy \textit{Highway policy} that works as an express train for the agent to go to the landmarks.

To find landmarks that can represent an area of the given distribution over states, we learn the landmark encoder $\texttt{enc}_\phi(s)$ and decoder $\texttt{dec}_\phi(l)$ through VQ-VAE.
We aim to encode model states $s$ into the $N$ learnable codes which we call landmark $l$ of a codebook, and vice versa. 

We encode the model states into embeddings using landmark encoder $\texttt{enc}_{\phi}(s)$. For quantization, the embedding $\texttt{enc}_{\phi}(s)$ is assigned to the closest code in the codebook $l_k$ where $k = \argmin_{j}\Vert \texttt{enc}_{\phi}(s) - l_j \Vert_2, k \in 1 \cdots N$. With the landmark decoder $\texttt{dec}_{\phi}(l_k)$ , $l_k$ can be decoded back to state $s$.
The training objective is 
\begin{align}
L_l = \Vert s - \texttt{dec}_{\phi}(l_k)\Vert_2^2 + \beta \Vert \texttt{sg}(l_k) - \texttt{enc}_{\phi}(s)\Vert_2^2,
\end{align}
where $\texttt{sg(·)}$ denotes stop gradient. The loss is composed of reconstruction error of the decoder and commitment loss, which is the difference between embedded vectors and the codes in the codebook. The balance in the loss is managed by the hyperparameter $\beta$.
We assign only a single code in the codebook to each model state. Thus, landmarks can be seen as cluster centers partitioning the model states into the number of codes in the codebook \cite{mazzaglia2022choreographer,campos2020edl}.

\subsection{Building Blocks for Strategy}
\label{sec:building_blocks}
\textbf{Highway policy.} We train a landmark-conditioned policy $\pi_{l}(a_t|s_t, l)$ called \textit{Highway policy} through imagined trajectories.
\arxcg{G}iven a target landmark $l$, the objective of this policy is to reach the state of the target landmark $\hat{s}_l=\texttt{dec}_\phi(l)$.
To train the Highway policy, we design the reward with two terms:
\vspace{-.5em}
\begin{align}
\hspace{-.25cm}
r_{l}(s_t, l) = - \Vert \texttt{dec}_\phi(l) - s_t \Vert_2^2 + \sum_{i=1}^{K} \text{log} \Vert s_t - s_i^{\text{K-NN}} \Vert_2
\end{align}
\vspace{-.2em}
The first term calculates the distance in the state space between the visited state $s_t$ and the decoded state from the conditioned landmark code $l$. This encourages the agent to reach the decoded state of $l$. The second term is estimated using a K-NN particle-based estimator \cite{singh2003nearest}, which motivates the agent to visit \arxcg{diverse} states within one trajectory. 


%


\textbf{Explorer and Achiever.} 
We follow prior approaches based on goal-conditioned MBRL framework \cite{mendonca2021lexa}.
\textit{Explorer} is an exploration policy $\pi_{e}(a_t|s_t)$ trained by receiving exploration reward $r_{e}(s_t)$. \arxcg{$r_{e}(s_t)$ encourages the policy to maximize the disagreement among an ensemble of 1-step dynamics models} \cite{pathak2019disagreement, sekar2020planning}. \arxcg{As the explorer trains in imagination, we start the imagined trajectories not only from sampled data from the replay buffer but also from landmarks.} We call the goal-conditioned policy $\pi_{g}(a_t|s_t, e_g)$ \textit{Achiever} that receives current model state and goal embedding $e_g = \texttt{enc}_\theta(x_g)$ as inputs, where $x_g$ is \arxcg{the goal image}. The reward for reaching a goal $r_{g}(\hat{e}_t, e_g)$ is based on a self-supervised objective that focuses on the temporal distance that follows prior works \cite{mendonca2021lexa}, where it encourages the policy to reduce the number of actions needed to move from the current state to the goal state. \arxcg{$\hat{e}_t = \texttt{emb}(s_t) \approx e_t $ is the predicted image embedding at step $t$} \arxcg{(refer to Appendix \ref{appendix:implementation} for more details)}.

\subsection{Strategy to Explore}
\label{sec:discovery}
\textit{How can the generalist agent strategically dream to explore during training time so that it can achieve diverse goals?}
Prior works leverage the world model for planning from randomly sampled candidate states \cite{mendonca2021lexa}.
However, in large or complex search spaces, chances of stumbling upon good solutions by random sampling are typically low \cite{ecoffet2021goexplore}. This leads to a lot of computational resources wasted on exploring sub-optimal areas.
Instead, we propose to plan strategically through dreaming by only evaluating the landmarks. 
By constructing the landmarks to represent the agent's experience (divide) and evaluating (conquer) only the representations of the explored space, we can gain a comprehensive approximation with efficiency. \arxcg{We also refer to this strategy as ``strategic exploration."}

We call the landmark with the highest exploration potential \textit{``Curious Landmark"}. We then move to the \textit{curious landmark} via \textit{Highway policy}, then resume to explore immediately with \textit{Explorer}.

\textbf{Curious landmark} should lead us to effective exploration in the future, entailing high future exploration reward potential. 
To select a curious landmark, we get the decoded model state $s_0^{(i)} \sim \texttt{dec}_\phi(l_i), i \in 1\hdots N$ of each landmark via landmark decoder. 
We then imagine $H$ steps trajectories with the Explorer through world model from each landmark, $\tau_i = \{s_0^{(i)}, s_1^{(i)},\hdots, s_H^{(i)}\}$. \arxcg{We calculate the curiosity $C_i$ of landmark $l_i$ as the expected exploration reward of $\tau_i$:}
%
\begin{align}
C_i = \mathbb{E}_{\tau_i}[r_{e}(s_t^{(i)})], \quad \tau_i = \{s_0^{(i)}, s_1^{(i)},\hdots, s_\arxcg{H}^{(i)}\}
\end{align}
\arxcg{Such that $r_e$ represents the exploration reward, as previously mentioned.} We then \arxcg{sample} the Curious Landmark $l_C$ with the probability of $C_i$. \arxcg{The curiosities of the landmarks are updated during the explorer's training.} 

Note that we are evaluating discrete states (landmarks), each playing a role as cluster centers dividing the model states into $N$ partitions. This enables us to have a comprehensive evaluation of the covered space efficiently, and \textit{Dream Strategically} takes advantage of the divide-and-conquer-like approach.

\textbf{Landmark-guided Exploration.} During exploration, we iterate over three phases: Every iteration starts with selecting a Curious Landmark $l_C$. Then, we exploit the \textit{Highway policy} $\pi_{l}(a_t|s_t, l_C)$ in the environment to reach $l_C$. If the Highway policy has been running for more than $T_{L}$ steps, Explorer takes over immediately and starts to explore. However, if the current state $s_t$ is near enough $s_{l_C} \sim \texttt{dec}_\phi(l_C)$
where the difference is under a certain threshold before $T_{L}$,
the agent switches to Explorer as well.

Explorer can start from a position with high exploration potential right away, reducing the time to visit previously well-known places and collecting high exploration value trajectories.
The iteration is repeated every $T_F$ step, maintaining a hierarchical structure.

\subsection{Strategy to Achieve}
\label{sec:conquer}
\textit{How can the agent efficiently train to reach numerous user-defined goals at test time? Is there a way to exploit the divide-and-conquer manner of \textit{strategically dreaming} at test time?}
We introduce the divide-and-conquer strategy once again, by finding the landmark that is nearest to the given goal and utilizing the Highway policy to reach the area closest to the goal (divide). Only then we exploit a local goal-conditioned policy trained to reach between close states (conquer). We call this goal-conditioned policy ``Achiever with \textit{Focused Sampling}", where it is trained to move between nearby states, thereby precisely mastering local areas. \arxcg{We demonstrate that by leveraging the divide-and-conquer strategy, we can achieve increased accuracy. We also refer to this strategy as ``strategic achievement."}

\textbf{Focused Sampling.}
Through the divide-and-conquer strategy, the Achiever $\pi_{g}(a_t|s_{t}, e_{g})$ is expected to be positioned very close to the goal when the policy is triggered. Thus, it only needs to cover a very short distance to reach its destination.
Instead of sampling random states from past trajectories like prior work \arxcg{\cite{mendonca2021lexa}}, we sample two different observations $x_{t}$, $x_{t+k}$ within the range $T_S$ in the same trajectory from the replay buffer. We use them as a starting state and goal state to train the Achiever, where $s_{t}$ is estimated through the world model from $x_t$ and $e_{g}$ is computed through the world model encoder $\texttt{enc}_\theta (x_{t+k})$.

Through this sampling, the policy is trained for the agent to navigate between states that are in close proximity, thereby improved sample efficiency is expected while exploiting the divide-and-conquer strategy to the full extent. We empirically investigate the efficacy of the focused sampling in our ablation study in Section \ref{sec:ablation_studies}.

\textbf{Landmark-guided Achievement.} At test time, we receive the user-defined pixel-level goal $x_g$ and estimate $s_g$ through the world model.
The agent estimates the landmark $l_G$ nearest to the goal state $s_g$, where $l_G = \argmin_j \Vert \texttt{enc}_\phi (s_g) - l_j \Vert _2$. We then utilize Highway policy $\pi_g(a_t|s_t, l_G)$ conditioned on $l_G$.
We switch to the Achiever $\pi_{g}(a_t|s_t, e_g)$ to reach the final goal when the highway policy has been running for more than $T_L$ steps, or when the current state is near enough to the landmark similar to landmark-guided exploration in Section~\ref{sec:discovery}).

We exploit Highway policy to move long distances conditioned on a small number of discrete landmarks, then utilize Achiever specialized to achieve nearby destinations, thereby achieving precision and scalability at the same time.
\section{Experiments}
\label{sec:experiments}
This section aims to evaluate the proposed agent by addressing the following questions: (1) Does Dr.~Strategy demonstrate improved performance than prior goal-conditioned MBRL works in zero-shot adaptation? (2) What is the role of the ``Strategy to Explore'' in enhancing exploration? (3) How does ``Strategy to Achieve'' contribute to improving zero-shot performance? (4) Does ``focused sampling'' for training the Achiever improve zero-shot performance?

\begin{figure}
    \centering
    \includegraphics[width=1\linewidth,  trim={7mm 0mm 0 0},clip]{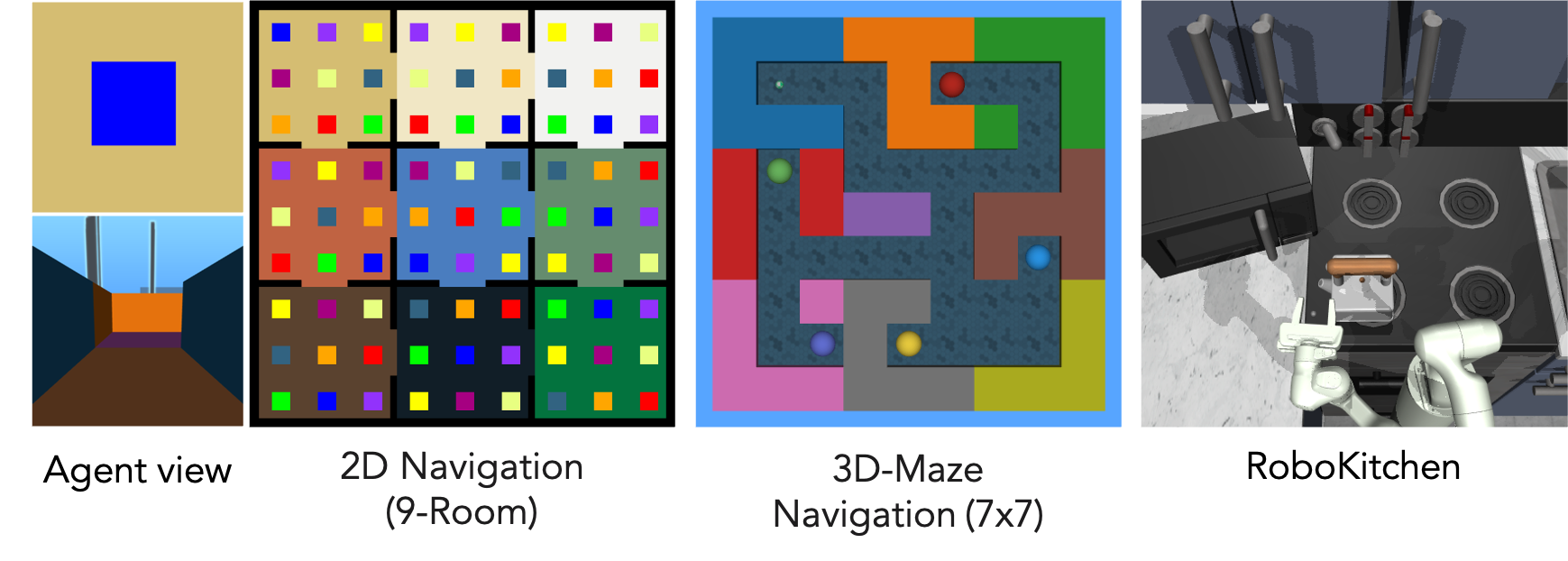}
    \vspace{-2em}
    \caption{
    \textbf{Environments.} \arxcg{We evaluate our agent across three different environments: 2D Navigation, 3D-Maze Navigation, and RoboKitchen. In these navigation environments, the agent's views are partially observable and visualized on the left. The top-left and bottom-left images represent the agent's initial view in the 2D and 3D Navigation settings, respectively. The second and third columns depict the top-down views of the 2D and 3D Navigation environments, respectively.}}
    \label{fig:environments}
    \vspace{-1.7em}
\end{figure}
\begin{figure*}
    \centering
    \includegraphics[width=\linewidth, trim={0 7mm 0 0},clip]{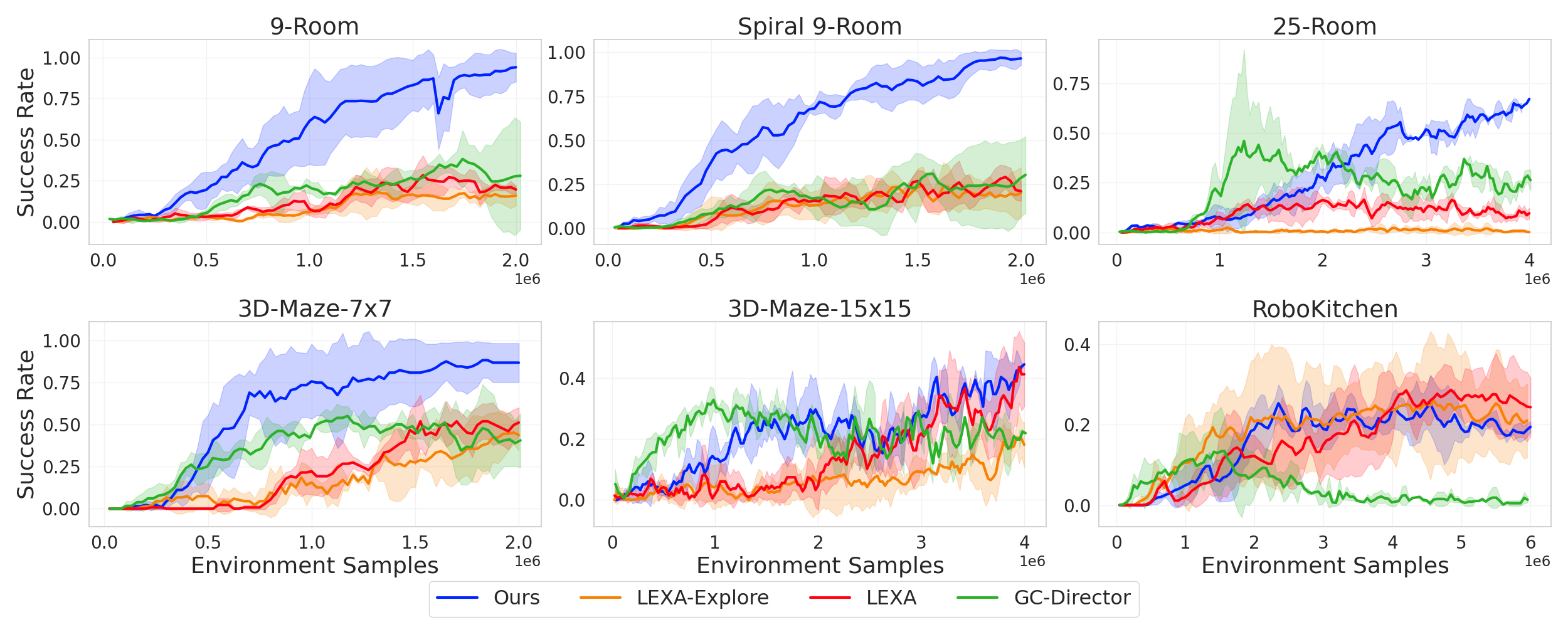}
    \vspace{-1em}
    \caption{\textbf{Zero-shot evaluation of the baselines across different environments.} Each baseline is evaluated given a goal image from the environment's test set. Dr. ~Strategy significantly outperforms other baselines in most of the navigation tasks, while achieving comparable results in RoboKitchen. The success rate is reported with the mean and standard deviation across 3 different random seeds.}
    \label{fig:main_results_success_rate}
    \vspace{-1em}
\end{figure*}

\subsection{Environments and Tasks}

To empirically investigate the proposed agent, we evaluate it in two types of navigation environments and a robot manipulation environment. One type of navigation environment is 2D navigation, in which the agent observes a partially observable limited top-down view as shown in Figure \ref{fig:environments}. We introduce three layouts: 9-room, 25-room, and spiral 9-room. The first two intend to test the agent's exploration capabilities in large spaces \cite{pertsch2020gcp}. 
The spiral 9-room layout (illustrated in Figure \ref{fig:appendix:envs_overview}) is specifically designed to challenge our agent's strategic exploration. It provides such a scenario where the exploration from the starting point can be inefficient due to the longer path to the farthest room \cite{ecoffet2021goexplore}.

We have designed a 3D-Maze navigation to evaluate the agent in a visually more complex environment, by modifying the Memory Maze environment \cite{pasukonis2022memmaze}. This provides the first-person view observation. We evaluate the agent's performance on two maze sizes: Maze-7x7 and Maze-15x15. 

Additionally, our evaluation extends to a robot manipulation environment, the RoboKitchen benchmark introduced in a prior work \cite{mendonca2021lexa}.
It features a third-person view of a 7-DoF Franka Emika Panda robotic arm equipped with a gripper. We note that it is a fully observable environment. The RoboKitchen environment requires the agent to interact with various objects, including microwave, kettle, light switch, burner, sliding cabinet, and hinge cabinet.
More details are discussed in Appendix~\ref{appendix:env}.

\subsection{Baselines}

We mainly compare Dr.~Strategy with \textbf{LEXA} \cite{mendonca2021lexa} because it is the closest model to ours but without the concept of strategic dreaming. It is also the state-of-the-art unsupervised model-based generalist agent for pixel-based observation tasks. In LEXA, the dreaming or imagination is guided by a rather naive strategy, i.e., random sampling from the replay buffer.

Regarding LEXA, it has been shown that only using the Explorer for the interaction can be better in a prior work \cite{hu2023peg}. Thus, we also test this baseline named \textbf{LEXA-Explore}.
LEXA, LEXA-Explore, and our model all share a similar high-level component structure in implementation. For fair comparison to minimize the effect of implementation engineering, we implemented LEXA and LEXA-Explore based on our Dr.~Strategy codebase.
The comparison with the original code can still be found in Appendix~\ref{appendix:baselines} and Appendix~\ref{appendix:LEXA-Original}.

Director \cite{hafner2022director} is a hierarchical model-based agent where a high-level policy (known as the manager) provides sub-goals to a low-level policy (known as the worker) to achieve a task defined by a reward function.
We chose Director due to its hierarchical structure and use of sub-goals, which is similar to exploiting landmarks in Dr.~Strategy. However, since Director is a task-specific agent \arxcg{and not a goal-conditioned agent}, it lacks the ability to generalize to diverse goals not given during training. Thus, we develop a goal-conditioned version of Director, named \textbf{GC-Director}. GC-Director utilizes a form of structure in the state space to achieve the given goal. The implementation details are discussed in Appendix \ref{appendix:baselines}.

\subsection{Main Results}

We conduct comparative analyses of our proposed agent with baselines across the three environments. The zero-shot evaluation performance is illustrated in Figure \ref{fig:main_results_success_rate}. These results are quantified based on the agent's success rate, which is determined by the distance to the goal. It is considered successful when the distance falls below a certain threshold (refer to Appendix~\ref{appendix:implementation} for more details).
We note that the goal images are unseen during training and are user-defined during test time, and the agent has to reach there. 

\begin{table}[hb]
    \centering
    \small 
    \begin{tabular}{@{}lccc@{}}
        \toprule
        \textbf{Method} & \textbf{9-Room} & \textbf{Spiral 9-Room} & \textbf{25-Room} \\
        \midrule
        LEXA & 19.75\% & 21.19\% & 9.62\% \\
        LEXA-Explore & 16.04\% & 20.16\% & 0.14\% \\
        GC-Director & 28.08\% & 30.45\% & 27.11\% \\
        \toprule
        \textbf{Dr. Strategy (Ours)} & \textbf{94.03\%} & \textbf{96.50\%} & \textbf{67.11\%} \\
        \bottomrule
    \end{tabular}
    \caption{Final success rate in 2D Navigation tasks.}
    \label{fig:miniworld_success_table}
\end{table}

\textbf{2D Navigation} 
In Table~\ref{fig:miniworld_success_table}, Dr.~Strategy shows an almost 100\% success rate in 9-room and spiral 9-room after 2M interaction steps with a clear performance gap compared to other baselines.
It is notable that our agent maintains a high success rate in spiral 9-room where the map is complicated and requires a longer range of exploration to achieve the goals in the farthest room.
A similar trend is observed in the 25-room layout, where the performance decreased to 67.11\%, as shown in Table~\ref{fig:miniworld_success_table}. However, the performance gap here is over 40\% compared to other baselines.
Interestingly, as the layout size increases, the agents with non-strategic achievement, LEXA and LEXA-Explore show a more significant performance deterioration. Conversely, GC-Director shows less performance decrease than others.
This result suggests that our strategic dreaming is more effective compared to the naive dreaming of LEXA or LEXA-Explore. 

\begin{table}[t!]
    \centering
    \small 
    \begin{tabular}{@{}lccc@{}}
        \toprule
        \textbf{Method} & \textbf{Maze-7x7} & \textbf{Maze-15x15} & \textbf{RoboKitchen} \\
        \midrule
        LEXA & 51.11\% & 41.20\% & \textbf{24.30}\% \\
        LEXA-Explore & 43.70\% & 18.05\% & 20.07\% \\
        GC-Director & 40.55\% & 21.87\% & 1.45\% \\
        \toprule
        \textbf{Dr. Strategy (Ours)} & \textbf{86.66\%} & \textbf{44.44\%} & 19.44\% \\
        \bottomrule
    \end{tabular}
    \vspace{-1.2em}
    \caption{Final success rate in 3D-Maze navigation and RoboKitchen tasks.}
    \label{fig:mzkitchen_success_table}
    \vspace{-2.1em}
\end{table}

\textbf{3D-Maze Navigation} 
For the Maze-7x7 environment providing visually more complex first-person observation, our agent achieves above 80\% success rate (refer to Table~\ref{fig:mzkitchen_success_table}) and significantly outperforms the baselines. Interestingly, all baselines, LEXA, LEXA-Explore, and GC-Director show better performances than they did for the 2D navigation environments. 
This may be due to the fact that the size of 3D-Maze Navigation maps is smaller than 2D Navigation maps: Maze-7x7 is about the size of four rooms in 2D Navigation maps and also has narrower corridors (as shown in Figure \ref{fig:environments}). This reduces the number of places the agent has to visit. 
With a smaller exploration space, this could be beneficial for baselines without strategic dreaming, leading to a smaller performance gap with Dr.~Strategy. 
However, despite such factors, Dr.~Strategy outperforms the baselines.
In Maze-15x15, our proposed agent outperforms the baselines yet, but the performance gap is reduced. It is because larger regions are identified with the same colors (illustrated in Figure \ref{fig:appendix:envs_overview}), which causes confusion for the highway policy to identify the landmark positions.

\textbf{RoboKitchen} 
The results are shown in Figure \ref{fig:main_results_success_rate} and Table~\ref{fig:mzkitchen_success_table}. Dr.~Strategy shows comparable performance with LEXA and LEXA-Explore, while GC-Director shows much worse performance than other agents. This is likely because of the environment's stationary view given in the third-person point, which decreases the visual distinctions between time steps. This can be critical in forming diverse and distinguishable landmarks based on reconstruction rewards. Furthermore, RoboKitchen tasks requires a short span of actions to achieve the goals compared to navigation tasks, which may reduce the need for strategic dreaming compared to other tasks. This is also supported by the low success rate of GC-Director, which also utilizes a hierarchical structure which is beneficial for long-horizon tasks.

\vspace{-.1cm}
\subsection{Qualitative Results}
\begin{figure}[!t]
\centering 
\begin{subfigure}{\linewidth}
   \centering
    \includegraphics[width=0.8\linewidth]{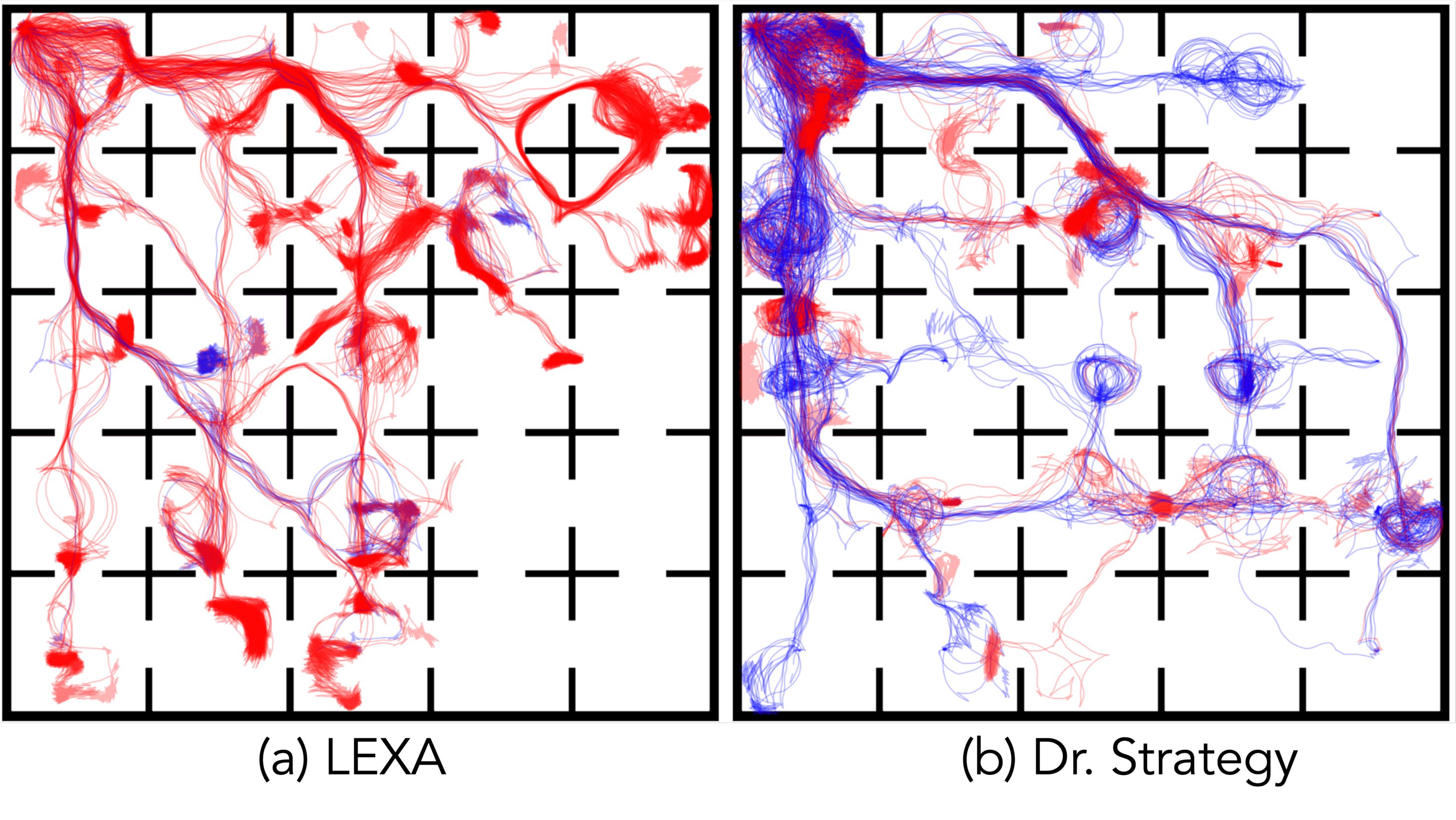}
    \vspace{-1.4em}
\end{subfigure}
\begin{subfigure}{\linewidth}
   \centering
    \includegraphics[width=0.82\linewidth]{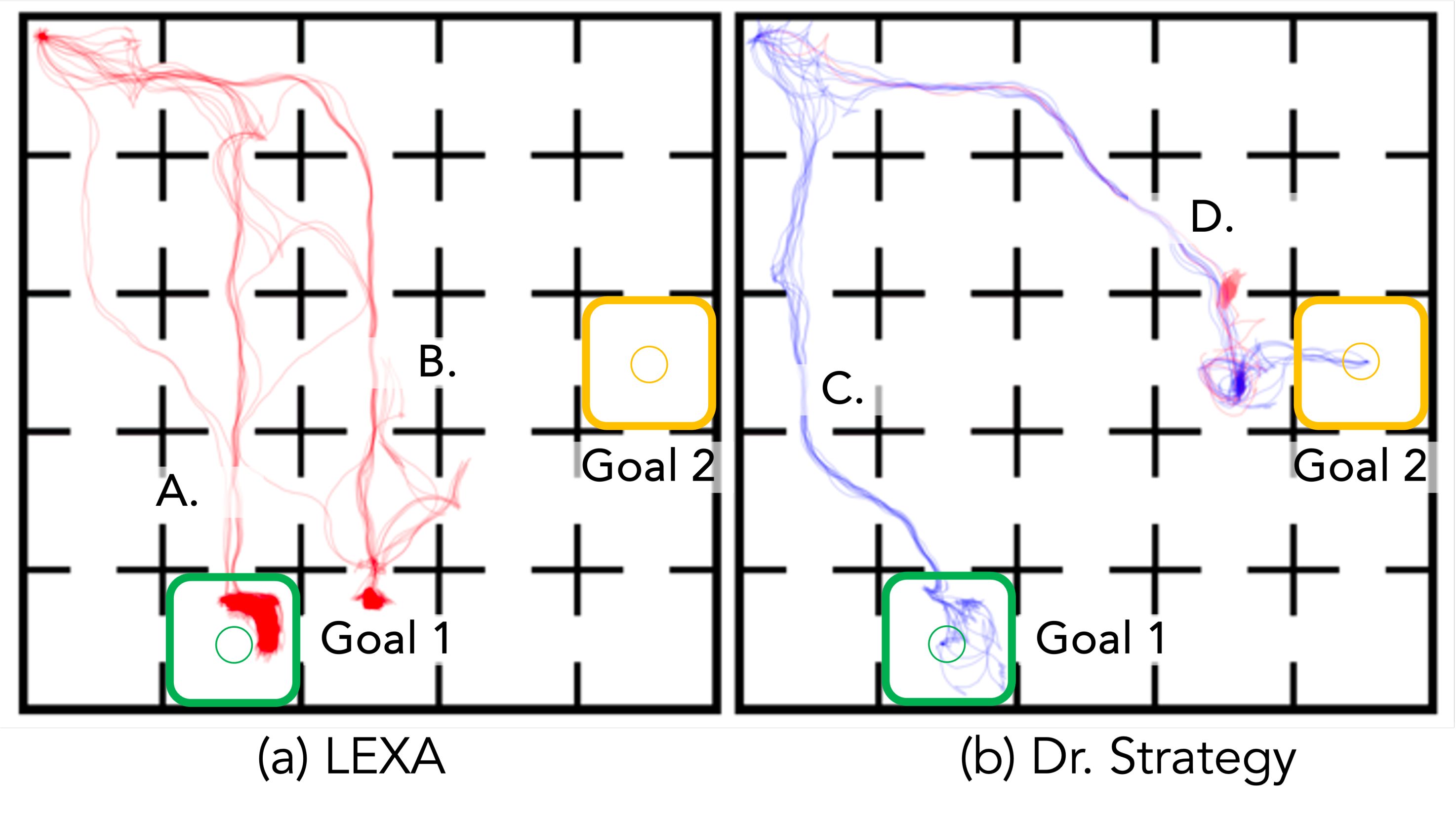}
   \vspace{-1.5em}
\end{subfigure}
\caption{\textbf{Evaluation trajectories visualization in 25-room for Dr.~Strategy and LEXA.} \arxcg{\textbf{(Top)} Ten evaluation trajectories per goal are visualized. All trajectories start from the top-left cell and head towards the desired goals positioned in the middle of each room. The \red{\textbf{red}} and \blue{\textbf{blue}} lines indicate failed and successful trajectories, respectively. \textbf{(Bottom)}} Trajectories (A), (C) aim to reach \textit{Goal 1} while (B), (D) aim to reach \textit{Goal 2}. Dr.~Strategy's trajectory (C) successfully reaches \textit{Goal 1} with precision due to focused sampling, unlike LEXA's trajectory (A). For \textit{Goal 2}, trajectory (D) demonstrates the advantages of exploiting highway policy by finding the goal's vicinity, a capability lacking in trajectory (B) with flat models. \comment{Overall, we can verify the performance gap between LEXA and Dr.~Strategy and confirm that Dr.~Strategy is able to reach more diverse, further goals. }}
\vspace{-.5em}
\label{fig:qualitative_trajectories}
\end{figure}

To investigate more details of the improvement through the strategic imagination, we visualize the trajectories of our proposed agent and LEXA on the 25-room layout in the 2D Navigation environment in Figure~\ref{fig:qualitative_trajectories}. We find that our agent can reach more diverse, further goals with higher success rates.
Moreover, we can examine the failures of LEXA: Both trajectories (A) and (C), highlighted as green boxes, aim to acquire goal 1. However, while trajectory (C) is able to reach the goal with high accuracy, trajectory (A) fumbles around the goal in close but not precise positions. It is because Dr.~Strategy learns to achieve with high precision through \textit{focused sampling}. This highlights the benefit of localizing the scope of the Achiever via the divide-and-conquer strategy.
Meanwhile, where both trajectories (B) and (D) aim to reach goal 2. However, trajectory (B) cannot even go near the desired goal 2. This shows the benefit of strategic dreaming, where the agent can find and move to the nearby area of goals, while flat models cannot plan such structured navigation and cannot locate near areas once it is lost.

\subsection{Ablation Studies} \label{sec:ablation_studies}

We investigate the influence of three components of strategic dreaming: Strategy to Explore (Section \ref{sec:discovery}), Strategy to Achieve, and focused sampling (Section \ref{sec:conquer}).

\textbf{Strategy to Explore.} To investigate the efficacy of strategic exploration, we compared the Dr.~Strategy with and without strategic exploration. The Dr.~Strategy without strategic exploration explores the environment similar to LEXA or LEXA-Explore \cite{mendonca2021lexa, hu2023peg}. 

\begin{figure}
    \centering
    \includegraphics[width=\linewidth, trim={0 2mm 0 0},clip]{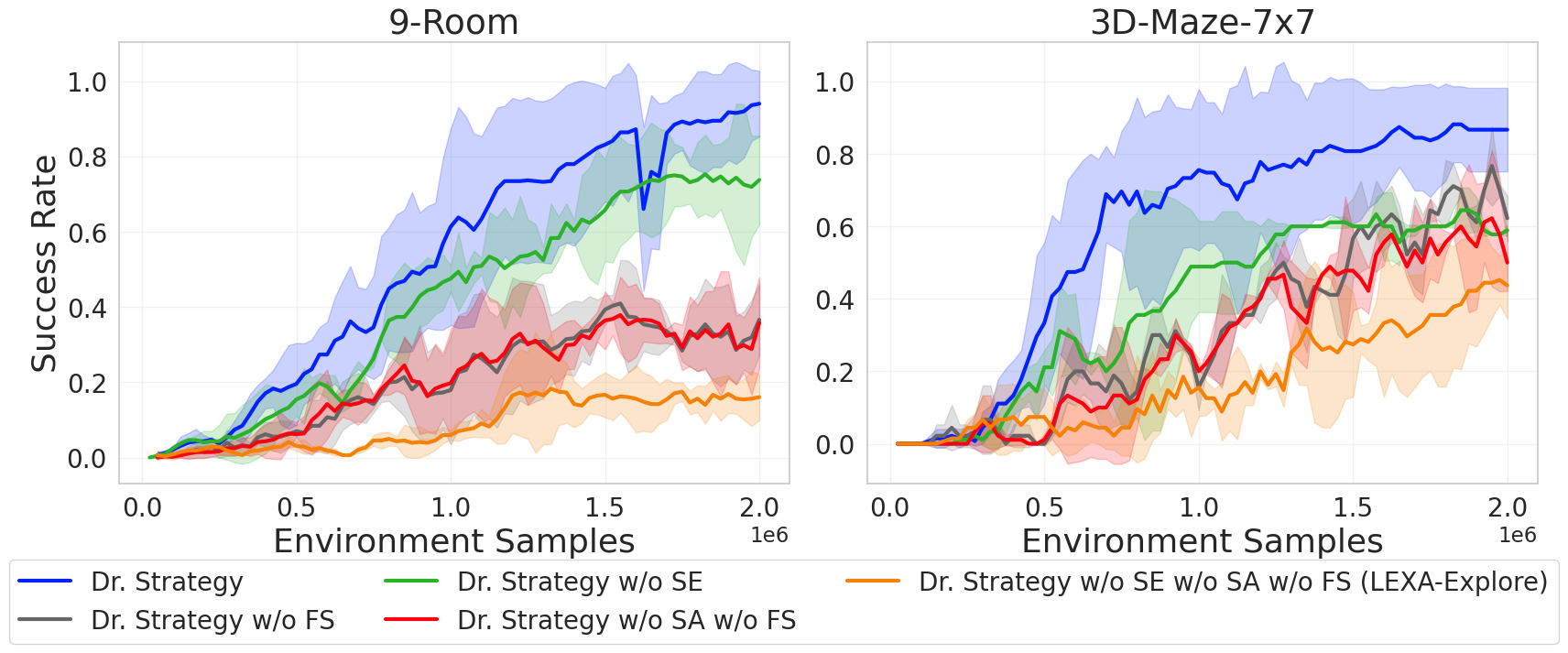}
    \vspace{-2em}
    \caption{\textbf{Ablation results for SE, SA, FS.} showing the influence of using Strategy to Explore (SE), Strategy to Achieve (SA), and focused sampling (FS) to Dr. ~Strategy's zero-shot success rate}
    \label{fig:ablation_modules}
    \vspace{-1.5em}
\end{figure}

In Figure \ref{fig:ablation_modules}, we compare the variants of Dr.~Strategy when the strategic exploration is applied and not in the aspect of the unseen goal achievement success rates. When comparing the variants with and without strategic exploration, we can find a clear performance gap regardless of equipping the strategic achievement and focused sampling. 

\textbf{Strategy to Achieve.} We also hypothesized that strategic achievement could be crucial to improving the unseen goal achievement performance. To study this, we compare Dr.~Strategy with and without strategic achievement. We note that the ablation version does not utilize the focused sampling, because the sampling is designed for strategic achievement. The result is shown in Figure \ref{fig:ablation_modules}.
The performance gaps between with and without strategic achievement are clearly shown regardless of strategic exploration. Its performance gap is larger than the gap from the ablation study for strategic exploration especially for 9-room, where we can find that the major performance gain of our agent compared to the naive dreaming versions such as LEXA or LEXA-Explore happened through this strategic achievement.
We note that we do not compare our agent with another model-based generalist agent PEG \cite{hu2023peg} that equips strategic exploration because it is designed for state-based environments. However, this result suggests that strategic exploration is not efficient enough, and the agent with strategic exploration and achievement (our agent) outperforms the agent only with strategic exploration such as PEG, and the agent without the strategic approach like us (LEXA).

\textbf{Focused Sampling.} To enhance achievement through the divide-conquer approach, we utilize focused sampling (Discussed in Section \ref{sec:conquer}) for the Achiever to achieve near goals. It is expected to improve the sample efficiency in Achiever training while fitting in the divide-conquer approach scenario. We study the expected efficacy by comparing it with the Dr.~Strategy without focused sampling. It is shown in Figure \ref{fig:ablation_modules} (compared with Dr.~Strategy without Focused Sampling (FS)). Surprisingly, the performance gap is huge, and without focused sampling, the agent performance is similar to the ablation without strategic achievement and focused sampling. This result suggests that training the Achiever with the nearby goals from the starting point is crucial to improve the performance and it can be available through the strategic achievement with the latent landmarks and highway policy.
\section{Related Work}

As an unsupervised model-based generalist agent, LEXA \cite{mendonca2021lexa} and PEG \cite{hu2023peg} are related to our work. However, LEXA is trained with naive strategic dreaming, which limits its performance in small size of state space \cite{hu2023peg}. PEG extends LEXA to apply the strategic exploration by exploring from the samples estimated as interesting through the roll-out of the explorer in the imagination like us, but they did not validate their method to the pixel-based environment and extend this strategy to the achiever like us. In the aspect of training discrete representative states and policy in imagination, our work is related to Choreographer \cite{mazzaglia2022choreographer}. However, Chreographer fine-tuned the learned representative states and policy for the downstream task with a new hierarchical policy while ours is the unsupervised generalist agent. Director \cite{hafner2022director}, a hierarchical model-based agent can be related in the aspect of utilizing the intermediate state for solving the given task, but Director is designed for solving a single task, not the unsupervised generalist agent.

Dr.~Strategy explores the environment from the interesting spot called \textit{Curious landmark}. In \cite{ecoffet2021goexplore,recode,hu2023peg}, this idea has been studied to address the inefficiency when exploring from the starting point \cite{pathak2017curiosity, burda2018rnd, pathak2019disagreement, mazzaglia2022lbs}, while PEG \cite{hu2023peg} does not apply strategic achievement with this idea, and Go-Explore \cite{ecoffet2021goexplore} and RECODE \cite{recode} are model-free RL methods.

Our agent utilizes the goal-conditioned policies, the highway policy, and the achiever. The goal-conditioned policy has been studied to learn the trajectories in an unsupervised manner by sampling the goals from the data \cite{eysenbach2018diayn,yarats2021protorl,park2022lsd,park2024metra, mazzaglia2022choreographer, kim2023discodance}, or train the agent that can solve multiple tasks \cite{andrychowicz2017hindsight, eysenbach2019sorb, pong2019skew, pitis2020mega, mendonca2021lexa, hu2023peg, hafner2022director}. \arxcg{However, these methods do not utilize a goal-conditioned policy (i.e., the highway policy) combined with the exploration policy and the achiever policy to improve exploration quality and the achievement of unseen goals.}

%
\section{Conclusion}

In this paper, we propose Dr.~Strategy, a novel model-based strategic, general-purpose agent.
Inspired by the structured and strategic planning of humans, we designed this agent to utilize strategic dreaming for efficient exploration and goal achievement through planning.
To do this, the agent learns the latent landmarks representing their experience and three distinct policies: navigating to the landmarks (Highway policy), exploring from the landmarks (explorer), and achieving the given goal from the landmarks (achiever).
Different from the previous approaches \cite{mendonca2021lexa,hu2023peg}, by separating the roles of the policies strategically, our agent showed better performances in diverse complex and partial observable navigation environments.
Especially, the divide-and-conquer approach allows the achiever \arxcg{to} learn from nearby samples, which dramatically improves the performance of the agent. 

\textbf{Limitations and future work.} However, the agent has shown limited performance in a robotic manipulation environment.
The performance improvement in those environments could be a future work.
Additionally, the current agent treats the number of landmarks as a hyperparameter, but it would be interesting to make it gradually increase and adapt \cite{dp_mean}. 
Another promising direction could be the integration of a hierarchical framework within the highway policy to extend the agent's exploration and goal-achievement capabilities.


\section*{Impact Statement}

Strategic Dreaming, as implemented in the Dr. Strategy agent, represents a novel structure in model-based reinforcement learning, focusing on enhancing agents' planning capabilities to "dream" in a structured manner. This approach draws from cognitive science insights, employing a spatial divide-and-conquer strategy for problem-solving. In practical terms, Strategic Dreaming could revolutionize tasks that require complex spatial navigation and decision-making, such as urban planning, logistics, and autonomous vehicle routing. By enabling AI to efficiently learn and navigate through simulations, Strategic Dreaming can lead to more robust and reliable models that require less real-world data, thereby reducing the time and cost associated with training AI systems.

However, the implications of Strategic Dreaming extend beyond improved efficiency. As these agents become adept at navigating and planning in simulated environments, there is potential for them to supplant roles currently filled by humans, especially in fields that rely heavily on spatial and strategic planning. While this could lead to increased efficiency and safety, particularly in hazardous environments, it also raises societal and ethical questions about the displacement of jobs and the need for new frameworks to govern AI decision-making and accountability. However, such capabilites require more investigation and does not seem to be a near future. The development of Strategic Dreaming thus mandates a careful consideration of its societal impact, balancing the benefits of advanced navigation and planning capabilities with the ethical management of automation's societal effects.

\section*{Acknowledgements}
This work is supported by Brain Pool Plus Program (No. 2021H1D3A2A03103645) through the National Research Foundation of Korea (NRF) funded by the Ministry of Science and ICT and partly by the research support program of Samsung Advanced Institute of Technology.

\bibliography{references}

\begin{thebibliography}{44}
\providecommand{\natexlab}[1]{#1}
\providecommand{\url}[1]{\texttt{#1}}
\expandafter\ifx\csname urlstyle\endcsname\relax
  \providecommand{\doi}[1]{doi: #1}\else
  \providecommand{\doi}{doi: \begingroup \urlstyle{rm}\Url}\fi

\bibitem[Andrychowicz et~al.(2017)Andrychowicz, Wolski, Ray, Schneider, Fong, Welinder, McGrew, Tobin, Pieter~Abbeel, and Zaremba]{andrychowicz2017hindsight}
Andrychowicz, M., Wolski, F., Ray, A., Schneider, J., Fong, R., Welinder, P., McGrew, B., Tobin, J., Pieter~Abbeel, O., and Zaremba, W.
\newblock Hindsight experience replay.
\newblock \emph{Advances in neural information processing systems}, 30, 2017.

\bibitem[Berner et~al.(2019)Berner, Brockman, Chan, Cheung, D{\k{e}}biak, Dennison, Farhi, Fischer, Hashme, Hesse, et~al.]{openai_five}
Berner, C., Brockman, G., Chan, B., Cheung, V., D{\k{e}}biak, P., Dennison, C., Farhi, D., Fischer, Q., Hashme, S., Hesse, C., et~al.
\newblock Dota 2 with large scale deep reinforcement learning.
\newblock \emph{arXiv preprint arXiv:1912.06680}, 2019.

\bibitem[Burda et~al.(2019)Burda, Edwards, Storkey, and Klimov]{burda2018rnd}
Burda, Y., Edwards, H., Storkey, A., and Klimov, O.
\newblock Exploration by random network distillation.
\newblock In \emph{International Conference on Learning Representations}, 2019.

\bibitem[Campos et~al.(2020)Campos, Trott, Xiong, Socher, Gir{\'o}-i Nieto, and Torres]{campos2020edl}
Campos, V., Trott, A., Xiong, C., Socher, R., Gir{\'o}-i Nieto, X., and Torres, J.
\newblock Explore, discover and learn: Unsupervised discovery of state-covering skills.
\newblock In \emph{International Conference on Machine Learning}, pp.\  1317--1327. PMLR, 2020.

\bibitem[Cho et~al.(2014)Cho, Van~Merri{\"e}nboer, Gulcehre, Bahdanau, Bougares, Schwenk, and Bengio]{cho2014learning}
Cho, K., Van~Merri{\"e}nboer, B., Gulcehre, C., Bahdanau, D., Bougares, F., Schwenk, H., and Bengio, Y.
\newblock Learning phrase representations using rnn encoder-decoder for statistical machine translation.
\newblock \emph{arXiv preprint arXiv:1406.1078}, 2014.

\bibitem[Chun \& Jiang(1998)Chun and Jiang]{chun1998contextual}
Chun, M.~M. and Jiang, Y.
\newblock Contextual cueing: Implicit learning and memory of visual context guides spatial attention.
\newblock \emph{Cognitive psychology}, 36\penalty0 (1):\penalty0 28--71, 1998.

\bibitem[Ecoffet et~al.(2021)Ecoffet, Huizinga, Lehman, Stanley, and Clune]{ecoffet2021goexplore}
Ecoffet, A., Huizinga, J., Lehman, J., Stanley, K.~O., and Clune, J.
\newblock First return, then explore.
\newblock \emph{Nature}, 590\penalty0 (7847):\penalty0 580--586, 2021.

\bibitem[Eysenbach et~al.(2019{\natexlab{a}})Eysenbach, Gupta, Ibarz, and Levine]{eysenbach2018diayn}
Eysenbach, B., Gupta, A., Ibarz, J., and Levine, S.
\newblock Diversity is all you need: Learning skills without a reward function.
\newblock In \emph{International Conference on Learning Representations}, 2019{\natexlab{a}}.

\bibitem[Eysenbach et~al.(2019{\natexlab{b}})Eysenbach, Salakhutdinov, and Levine]{eysenbach2019sorb}
Eysenbach, B., Salakhutdinov, R., and Levine, S.
\newblock Search on the replay buffer: Bridging planning and rl.
\newblock \emph{Advances in Neural Information Processing Systems}, 2019{\natexlab{b}}.

\bibitem[Guss et~al.(2019)Guss, Houghton, Topin, Wang, Codel, Veloso, and Salakhutdinov]{minerl}
Guss, W.~H., Houghton, B., Topin, N., Wang, P., Codel, C., Veloso, M., and Salakhutdinov, R.
\newblock Minerl: A large-scale dataset of minecraft demonstrations.
\newblock \emph{arXiv preprint arXiv:1907.13440}, 2019.

\bibitem[Ha \& Schmidhuber(2018)Ha and Schmidhuber]{worldmodel}
Ha, D. and Schmidhuber, J.
\newblock Recurrent world models facilitate policy evolution.
\newblock \emph{Advances in neural information processing systems}, 31, 2018.

\bibitem[Hafner et~al.(2019{\natexlab{a}})Hafner, Lillicrap, Ba, and Norouzi]{hafner2019dreamerv1}
Hafner, D., Lillicrap, T., Ba, J., and Norouzi, M.
\newblock Dream to control: Learning behaviors by latent imagination.
\newblock In \emph{International Conference on Learning Representations}, 2019{\natexlab{a}}.

\bibitem[Hafner et~al.(2019{\natexlab{b}})Hafner, Lillicrap, Fischer, Villegas, Ha, Lee, and Davidson]{hafner2019rssm}
Hafner, D., Lillicrap, T., Fischer, I., Villegas, R., Ha, D., Lee, H., and Davidson, J.
\newblock Learning latent dynamics for planning from pixels.
\newblock In \emph{International conference on machine learning}, pp.\  2555--2565. PMLR, 2019{\natexlab{b}}.

\bibitem[Hafner et~al.(2020)Hafner, Lillicrap, Norouzi, and Ba]{hafner2020dreamerv2}
Hafner, D., Lillicrap, T.~P., Norouzi, M., and Ba, J.
\newblock Mastering atari with discrete world models.
\newblock In \emph{International Conference on Learning Representations}, 2020.

\bibitem[Hafner et~al.(2022)Hafner, Lee, Fischer, and Abbeel]{hafner2022director}
Hafner, D., Lee, K.-H., Fischer, I., and Abbeel, P.
\newblock Deep hierarchical planning from pixels.
\newblock \emph{Advances in Neural Information Processing Systems}, 35:\penalty0 26091--26104, 2022.

\bibitem[Hu et~al.(2023)Hu, Chang, Rybkin, and Jayaraman]{hu2023peg}
Hu, E.~S., Chang, R., Rybkin, O., and Jayaraman, D.
\newblock Planning goals for exploration.
\newblock In \emph{International Conference on Learning Representations}, 2023.

\bibitem[Kaiser et~al.(2018)Kaiser, Bengio, Roy, Vaswani, Parmar, Uszkoreit, and Shazeer]{kaiser2018fast}
Kaiser, L., Bengio, S., Roy, A., Vaswani, A., Parmar, N., Uszkoreit, J., and Shazeer, N.
\newblock Fast decoding in sequence models using discrete latent variables.
\newblock In \emph{International Conference on Machine Learning}, pp.\  2390--2399. PMLR, 2018.

\bibitem[Kim et~al.(2023)Kim, Lee, Lee, Hwang, Park, Min, and Choo]{kim2023discodance}
Kim, H., Lee, B., Lee, H., Hwang, D., Park, S., Min, K., and Choo, J.
\newblock Learning to discover skills through guidance.
\newblock In \emph{Advances in Neural Information Processing Systems}, 2023.

\bibitem[Kingma \& Ba(2014)Kingma and Ba]{kingma2014adam}
Kingma, D.~P. and Ba, J.
\newblock Adam: A method for stochastic optimization.
\newblock \emph{arXiv preprint arXiv:1412.6980}, 2014.

\bibitem[Kingma \& Welling(2013)Kingma and Welling]{kingma2013auto}
Kingma, D.~P. and Welling, M.
\newblock Auto-encoding variational bayes.
\newblock \emph{arXiv preprint arXiv:1312.6114}, 2013.

\bibitem[Kulis \& Jordan(2011)Kulis and Jordan]{dp_mean}
Kulis, B. and Jordan, M.~I.
\newblock Revisiting k-means: New algorithms via bayesian nonparametrics.
\newblock \emph{arXiv preprint arXiv:1111.0352}, 2011.

\bibitem[Mazzaglia et~al.(2022{\natexlab{a}})Mazzaglia, Catal, Verbelen, and Dhoedt]{mazzaglia2022lbs}
Mazzaglia, P., Catal, O., Verbelen, T., and Dhoedt, B.
\newblock Curiosity-driven exploration via latent bayesian surprise.
\newblock In \emph{Proceedings of the AAAI Conference on Artificial Intelligence}, volume~36, pp.\  7752--7760, 2022{\natexlab{a}}.

\bibitem[Mazzaglia et~al.(2022{\natexlab{b}})Mazzaglia, Verbelen, Dhoedt, Lacoste, and Rajeswar]{mazzaglia2022choreographer}
Mazzaglia, P., Verbelen, T., Dhoedt, B., Lacoste, A., and Rajeswar, S.
\newblock Choreographer: Learning and adapting skills in imagination.
\newblock In \emph{International Conference on Learning Representations}, 2022{\natexlab{b}}.

\bibitem[Mendonca et~al.(2021)Mendonca, Rybkin, Daniilidis, Hafner, and Pathak]{mendonca2021lexa}
Mendonca, R., Rybkin, O., Daniilidis, K., Hafner, D., and Pathak, D.
\newblock Discovering and achieving goals via world models.
\newblock \emph{Advances in Neural Information Processing Systems}, 34:\penalty0 24379--24391, 2021.

\bibitem[Mnih et~al.(2015)Mnih, Kavukcuoglu, Silver, Rusu, Veness, Bellemare, Graves, Riedmiller, Fidjeland, Ostrovski, et~al.]{dqn}
Mnih, V., Kavukcuoglu, K., Silver, D., Rusu, A.~A., Veness, J., Bellemare, M.~G., Graves, A., Riedmiller, M., Fidjeland, A.~K., Ostrovski, G., et~al.
\newblock Human-level control through deep reinforcement learning.
\newblock \emph{nature}, 518\penalty0 (7540):\penalty0 529--533, 2015.

\bibitem[Mnih et~al.(2016)Mnih, Badia, Mirza, Graves, Lillicrap, Harley, Silver, and Kavukcuoglu]{a3c}
Mnih, V., Badia, A.~P., Mirza, M., Graves, A., Lillicrap, T., Harley, T., Silver, D., and Kavukcuoglu, K.
\newblock Asynchronous methods for deep reinforcement learning.
\newblock In \emph{International conference on machine learning}, pp.\  1928--1937. PMLR, 2016.

\bibitem[Park et~al.(2022)Park, Choi, Kim, Lee, and Kim]{park2022lsd}
Park, S., Choi, J., Kim, J., Lee, H., and Kim, G.
\newblock Lipschitz-constrained unsupervised skill discovery.
\newblock In \emph{International Conference on Learning Representations}, 2022.

\bibitem[Park et~al.(2024)Park, Rybkin, and Levine]{park2024metra}
Park, S., Rybkin, O., and Levine, S.
\newblock {METRA}: Scalable unsupervised {RL} with metric-aware abstraction.
\newblock In \emph{International Conference on Learning Representations}, 2024.

\bibitem[Pasukonis et~al.(2022)Pasukonis, Lillicrap, and Hafner]{pasukonis2022memmaze}
Pasukonis, J., Lillicrap, T., and Hafner, D.
\newblock Evaluating long-term memory in 3d mazes.
\newblock \emph{arXiv preprint arXiv:2210.13383}, 2022.

\bibitem[Pathak et~al.(2017)Pathak, Agrawal, Efros, and Darrell]{pathak2017curiosity}
Pathak, D., Agrawal, P., Efros, A.~A., and Darrell, T.
\newblock Curiosity-driven exploration by self-supervised prediction.
\newblock In \emph{International conference on machine learning}, pp.\  2778--2787. PMLR, 2017.

\bibitem[Pathak et~al.(2019)Pathak, Gandhi, and Gupta]{pathak2019disagreement}
Pathak, D., Gandhi, D., and Gupta, A.
\newblock Self-supervised exploration via disagreement.
\newblock In \emph{International conference on machine learning}, pp.\  5062--5071. PMLR, 2019.

\bibitem[Pertsch et~al.(2020)Pertsch, Rybkin, Ebert, Zhou, Jayaraman, Finn, and Levine]{pertsch2020gcp}
Pertsch, K., Rybkin, O., Ebert, F., Zhou, S., Jayaraman, D., Finn, C., and Levine, S.
\newblock Long-horizon visual planning with goal-conditioned hierarchical predictors.
\newblock \emph{Advances in Neural Information Processing Systems}, 33:\penalty0 17321--17333, 2020.

\bibitem[Pitis et~al.(2020)Pitis, Chan, Zhao, Stadie, and Ba]{pitis2020mega}
Pitis, S., Chan, H., Zhao, S., Stadie, B., and Ba, J.
\newblock Maximum entropy gain exploration for long horizon multi-goal reinforcement learning.
\newblock In \emph{International Conference on Machine Learning}, pp.\  7750--7761. PMLR, 2020.

\bibitem[Pong et~al.(2020)Pong, Dalal, Lin, Nair, Bahl, and Levine]{pong2019skew}
Pong, V., Dalal, M., Lin, S., Nair, A., Bahl, S., and Levine, S.
\newblock Skew-fit: State-covering self-supervised reinforcement learning.
\newblock In \emph{International Conference on Machine Learning}, pp.\  7783--7792. PMLR, 2020.

\bibitem[Razavi et~al.(2019)Razavi, Van~den Oord, and Vinyals]{razavi2019generating_vqvae2}
Razavi, A., Van~den Oord, A., and Vinyals, O.
\newblock Generating diverse high-fidelity images with vq-vae-2.
\newblock \emph{Advances in neural information processing systems}, 32, 2019.

\bibitem[Rezende et~al.(2014)Rezende, Mohamed, and Wierstra]{rezende2014stochastic}
Rezende, D.~J., Mohamed, S., and Wierstra, D.
\newblock Stochastic backpropagation and approximate inference in deep generative models.
\newblock In \emph{International conference on machine learning}, pp.\  1278--1286. PMLR, 2014.

\bibitem[Saade et~al.(2023)Saade, Kapturowski, Calandriello, Blundell, Sprechmann, Sarra, Groth, Valko, and Piot]{recode}
Saade, A., Kapturowski, S., Calandriello, D., Blundell, C., Sprechmann, P., Sarra, L., Groth, O., Valko, M., and Piot, B.
\newblock Unlocking the power of representations in long-term novelty-based exploration.
\newblock \emph{arXiv preprint arXiv:2305.01521}, 2023.

\bibitem[Sekar et~al.(2020)Sekar, Rybkin, Daniilidis, Abbeel, Hafner, and Pathak]{sekar2020planning}
Sekar, R., Rybkin, O., Daniilidis, K., Abbeel, P., Hafner, D., and Pathak, D.
\newblock Planning to explore via self-supervised world models.
\newblock In \emph{International Conference on Machine Learning}, pp.\  8583--8592. PMLR, 2020.

\bibitem[Singh et~al.(2003)Singh, Misra, Hnizdo, Fedorowicz, and Demchuk]{singh2003nearest}
Singh, H., Misra, N., Hnizdo, V., Fedorowicz, A., and Demchuk, E.
\newblock Nearest neighbor estimates of entropy.
\newblock \emph{American journal of mathematical and management sciences}, 23\penalty0 (3-4):\penalty0 301--321, 2003.

\bibitem[Sutton(1991)]{sutton1991dyna}
Sutton, R.~S.
\newblock Dyna, an integrated architecture for learning, planning, and reacting.
\newblock \emph{ACM Sigart Bulletin}, 2\penalty0 (4):\penalty0 160--163, 1991.

\bibitem[Sutton \& Barto(2018)Sutton and Barto]{sutton_rl}
Sutton, R.~S. and Barto, A.~G.
\newblock \emph{Reinforcement learning: An introduction}.
\newblock MIT press, 2018.

\bibitem[Van Den~Oord et~al.(2017)Van Den~Oord, Vinyals, et~al.]{van2017vqvae1}
Van Den~Oord, A., Vinyals, O., et~al.
\newblock Neural discrete representation learning.
\newblock \emph{Advances in neural information processing systems}, 30, 2017.

\bibitem[Vinyals et~al.(2019)Vinyals, Babuschkin, Chung, Mathieu, Jaderberg, Czarnecki, Dudzik, Huang, Georgiev, Powell, et~al.]{dm_starcraft2}
Vinyals, O., Babuschkin, I., Chung, J., Mathieu, M., Jaderberg, M., Czarnecki, W.~M., Dudzik, A., Huang, A., Georgiev, P., Powell, R., et~al.
\newblock Alphastar: Mastering the real-time strategy game starcraft ii.
\newblock \emph{DeepMind blog}, 2:\penalty0 20, 2019.

\bibitem[Yarats et~al.(2021)Yarats, Fergus, Lazaric, and Pinto]{yarats2021protorl}
Yarats, D., Fergus, R., Lazaric, A., and Pinto, L.
\newblock Reinforcement learning with prototypical representations.
\newblock In \emph{International Conference on Machine Learning}, pp.\  11920--11931. PMLR, 2021.

\end{thebibliography}
\bibliographystyle{icml2024}

\appendix
\onecolumn
\section{Environment}
\label{appendix:env}
\vspace{1em}


\begin{figure}[!t]
\centering 
    \includegraphics[
      width=15cm,
      height=10cm,
      keepaspectratio]{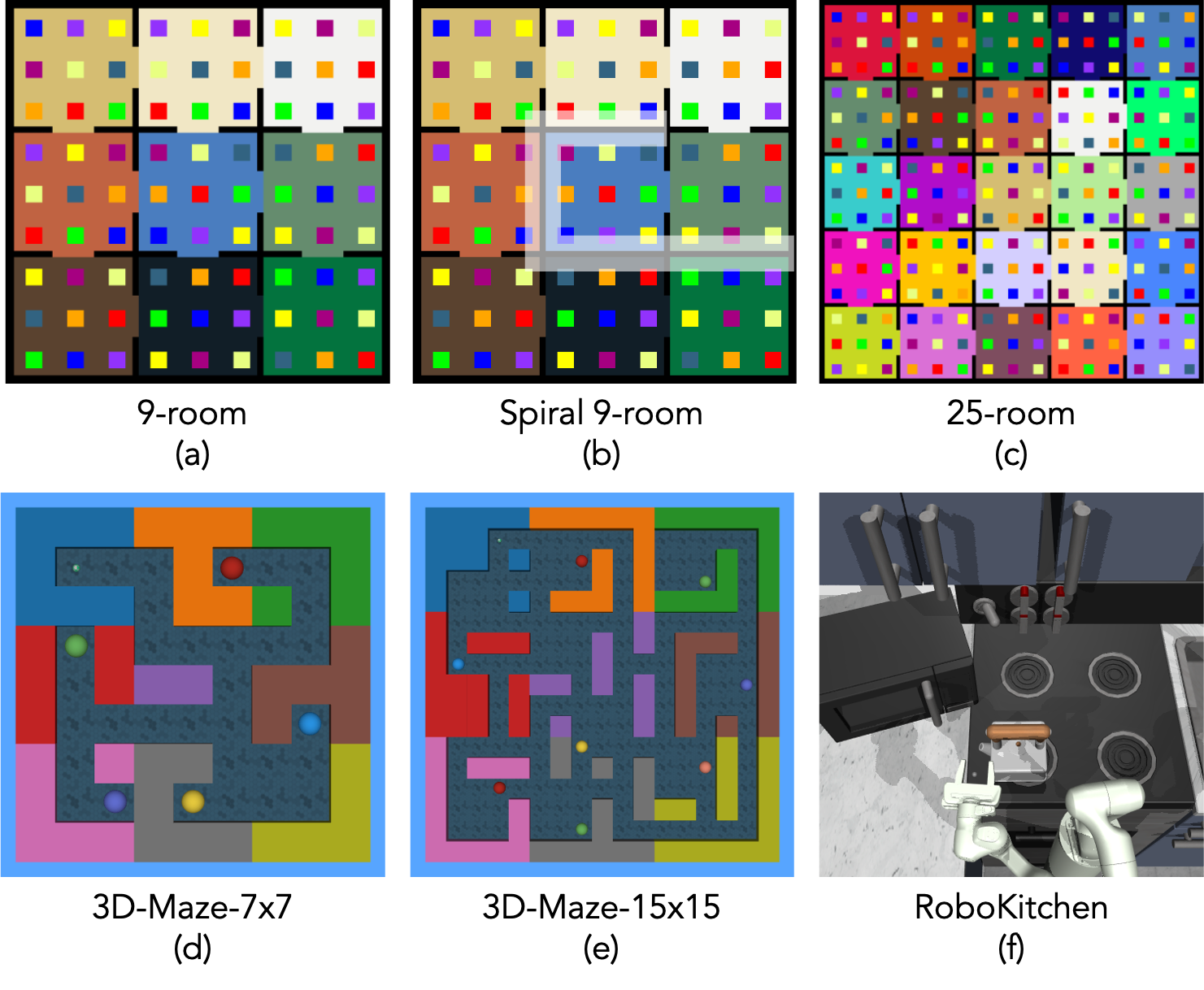}
    \vspace{-1em}
    \caption{\textbf{Illustration of all the used environments.} (a-c) Partially Observable 2D Navigation, (d-e) First-person view 3D maze navigation and (f) RoboKitchen. (b) shows the spiral 9-rooms in which the closed gates are highlighted in white, (d-e) showing the 3D-Maze environments without the floor color for easy visualizations of the walls}
    \label{fig:appendix:envs_overview}
\end{figure}
\vspace{1.5em}
\begin{figure}[!t]
   \centering
   \includegraphics[
      width=15cm,
      height=10cm,
      keepaspectratio]{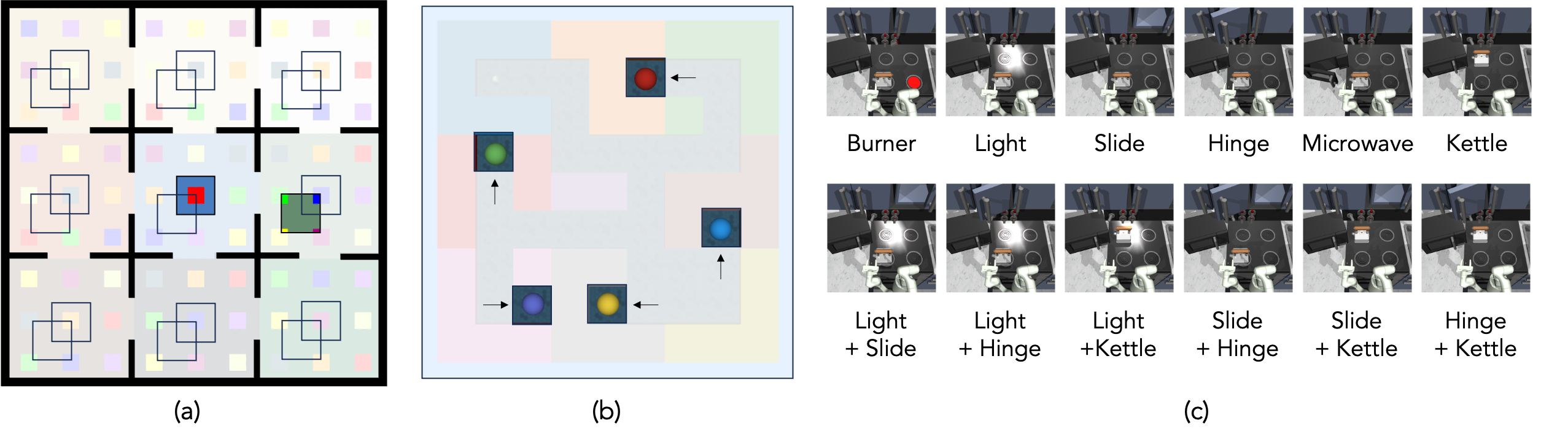}
    \vspace{-0.5em}
    \caption{\textbf{Zero-shot evaluation goals on each environment.} Our agent is evaluated given unseen goals in the evaluation phase. (a) and (b) illustrate the goals in navigation environments and (c) shows the goal images of the RoboKitchen benchmark.}
    \label{fig:envs_goals}
\end{figure}

\textbf{2D navigation.} We introduce three 2D navigation environments with distinct layouts: 9-room, spiral 9-room, and 25-room to evaluate the performance of structured and strategic imagination in large environments. All environments are modeled as egocentric views with limited visibility, represented by a 5x5 sized observation window as 64x64x3 pixel observation as shown in Figure \ref{fig:environments}. The agent aims to navigate through rooms of size 15x15 to reach specific points within a 0.1 Manhattan distance tolerance in 1000 steps. \arxcg{We calculate the agent's success rate per goal by averaging the outcomes of three evaluation episodes.} Each goal can be found at the center of a room or in the down-left corner of the 9-rooms and spiral 9-room layout and the center of a room in the 25-room layout as shown in Figure \ref{fig:envs_goals} (a). We note that these environments are non-episodic, requiring the agent to continuously explore and adapt without restarting episodes.

\textbf{3D-Maze.} We introduce two 3D-Maze navigation environments to assess the proficiency of structured and strategic imagination in navigating large and visually intricate spaces. We modified the Memory-Maze \cite{pasukonis2022memmaze} by keeping the same structure of the environment during all episodes to assist the agent in localization and distinguishing visual observation; we assign distinct colors to the walls and floors of the mazes. These mazes come in two sizes: 7x7 and 15x15, each designed with unique layouts as illustrated in Figure \ref{fig:appendix:envs_overview} (d) and (e). The environments are depicted from an egocentric viewpoint, limiting visibility to a 64x64x3 pixel observation as shown in Figure \ref{fig:environments}. The agent's objective is to reach specific points (illustrated in Figure \ref{fig:envs_goals} (b)) with a 0.1 Manhattan distance tolerance and 45 degrees of orientation tolerance, accomplishing this within 500 steps for the 7x7 maze and 1000 steps for the 15x15 maze. We measure whether the agent reached or not by three times and take an average to calculate the success rate per goal. The target points are strategically placed either at the dead ends of the maze or in proximity to the walls. Notably, these environments are non-episodic, requiring the agent to continually explore and adapt without restarting episodes.  

\textbf{Robokitchen.} To demonstrate the broad applicability of our agent, we chose the RoboKitchen environment from LEXA \cite{mendonca2021lexa} to evaluate its performance on robotic manipulation tasks requiring both structured and strategic imagination. We adopted the same setup as LEXA, setting the episode length to 150 steps with an action repeat factor of 2 with 12 visually distinguishable goals. We measure whether the agent reached or not by ten times and take an average to calculate the success rate per goal.

\begin{figure}[!t]
\centering 

   \includegraphics[width=15cm,
      height=10cm,
      keepaspectratio]{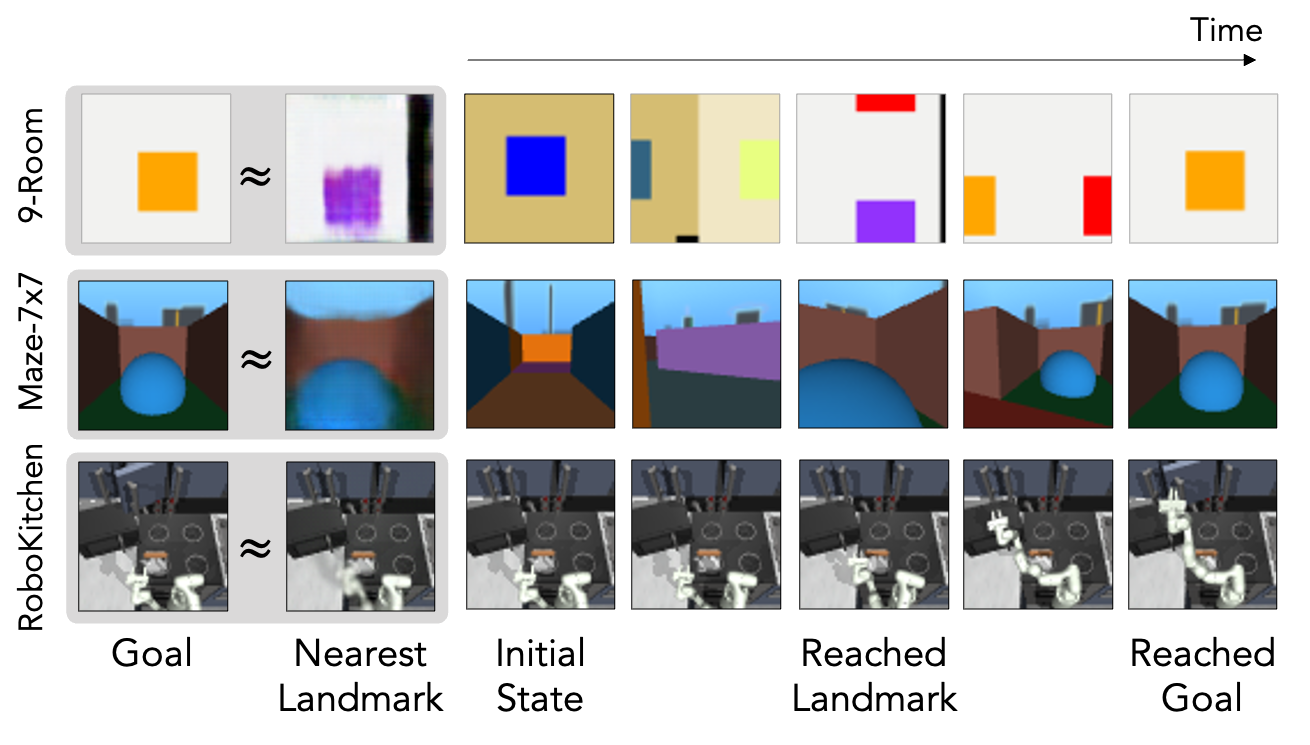}
    \caption{\textbf{Qualitative results of Dr.~Strategy's zero-shot evaluation trajectories.} Given the goal, the proposed agent finds the nearest landmark. We visualize it by inferring the latent state using the world model, and then it is reconstructed. The agent starts in the initial state and then uses the highway policy conditioned on the closest landmark. Upon meeting the termination criteria, it then switches to the focused achiever policy, conditioned on the given goal.}
    \label{fig:qualitative_reconstruction}
\end{figure}

\section{Baselines}
\label{appendix:baselines}

\arxcg{A primary approach in reinforcement learning (RL) to improve sample efficiency is via model-based reinforcement learning (MBRL) \cite{sutton1991dyna,worldmodel}. Dreamer \cite{hafner2019dreamerv1, hafner2020dreamerv2} is a MBRL agent that leverages the learning of an internal model, known as a world model (WM), to train an agent in dreaming also referred to as imagination. The world model is trained to predict the transition dynamics of the real environment. The agent trains in imagination via interacting with the WM instead of the real environment, facilitating faster experience collection for training. The collected trajectories via this interactions are called imagined trajectories. Thus, the world model serves as a proxy for the real environment. Dr.~Strategy and all baseline models employ Dreamer V2 \cite{hafner2020dreamerv2}, utilizing the world model for sample-efficient training.}

\textbf{LEXA} \arxcg{LEXA is a model-based RL agent that trains both an explorer and an achiever through imagination using a world model \cite{mendonca2021lexa}. The explorer discovers the environment, driven by intrinsic motivation, whereas the achiever gathers more experience by targeting randomly explored states sampled from the replay buffer. LEXA undergoes an unsupervised pre-training phase, after which the achiever attempts to solve tasks given by images in a zero-shot manner, without any further learning.} 
In comparison to the original LEXA setup, we opt for using disagreement \cite{pathak2019disagreement} as the intrinsic reward instead of latent disagreement \cite{sekar2020planning}. Moreover, our model incorporates a stochastic embedding sampled from a categorical one-hot distribution, akin to DreamerV2 \cite{hafner2020dreamerv2}, to modify the multi-diagonal Gaussian distribution.
This intentional variation in intrinsic rewards and sampling distributions aims to fine-tune performance specifically for 2D navigation environments.
For a fair comparison, we match LEXA's hyperparameters with our implementation, excluding latent landmark configurations as outlined in Appendix \ref{appendix:hyperparameters}. \arxcg{We reward the achiever policy for reaching the target state by using a temporal distance predictor, following the approach used in LEXA \cite{mendonca2021lexa}.}\comment{To assess whether the agent reaches the target state, we employ a temporal distance predictor, following the approach used in LEXA.}

\textbf{LEXA-Explore} \quad Building on PEG's \cite{hu2023peg} insight that excluding achiever-sampled trajectories \arxcg{benefits the success rate in LEXA}. Diverging from the original LEXA \cite{mendonca2021lexa}, we replaced latent disagreement \cite{sekar2020planning} with disagreement \cite{pathak2019disagreement} as an intrinsic reward. Furthermore, we adopted a stochastic embedding from a categorical one-hot distribution, akin to DreamerV2 \cite{hafner2020dreamerv2}, modifying the multi-diagonal Gaussian distribution. These adjustments aim to enhance performance in 2D navigation environments. Hyperparameters are matched with our method's implementation, excluding latent landmark configurations in line with Appendix \ref{appendix:hyperparameters}. \comment{To determine whether the agent reaches the target state, we employ a temporal distance predictor similar to LEXA.} 

\textbf{GC-Director} \quad Director \cite{hafner2022director} is a task-specific hierarchical model-based agent. The task is specified by the reward function. We develop GC-Director as a \textbf{G}oal-\arxcg{\textbf{C}}onditioned version of Director, to explore the environment and learn to achieve an unseen goal in an unsupervised manner similar to LEXA \cite{mendonca2021lexa}. 

Director includes two policies: high-level (manager), and low-level (worker). We developed GC-Director based on the open-source code of Director and followed the same architecture and training procedure of LEXA but using a hierarchical policy instead of the flat one in LEXA. GC-Director has 4 policies in total: Explorer has a manager and worker, and Achiever has \arxcg{another} manager and worker. We found that having two separate workers leads to the best results. 

The explorer's manager is rewarded by an intrinsic reward. The intrinsic reward is the estimate of the epistemic uncertainty using a disagreement of an ensemble of 1-step transition functions similar to LEXA's explorer. For the achiever, the manager $\pi_{mgr}^g(z\mid s_t, e_g)$ is conditioned on the embedding of the given goal image $e_g$ and is rewarded using the latent distance (either cosine similarity or temporal distance), as in LEXA. The worker in each is only trained using the original \arxcg{reward} function used in Director.

In Table \ref{tab:methods_diff}, we show a summary of the main aspects of the baselines. We denote hierarchical exploration by methods that have multiple policies that are used to explore sequentially and similarly for hierarchical achievement.

\begin{table}[htb]
   \vspace{-.5em}
   \small
   \centering
   \resizebox{.8\columnwidth}{!}{

\begin{tabular}{
  l
  >{\centering\arraybackslash}p{1.8cm}
  >{\centering\arraybackslash}p{1.8cm}
  >{\centering\arraybackslash}p{1.8cm}
  >{\centering\arraybackslash}p{1.8cm}}
  \toprule
    &  \textbf{Goal-Conditioned} &  \textbf{Hierarchical Exploration} &  \textbf{Hierarchical Achievement} & \textbf{Strategic Dreaming} \\
   \midrule
   LEXA \cite{mendonca2021lexa} &  \textcolor{green}{\ding{51}} &  \textcolor{red}{\ding{55}} &  \textcolor{red}{\ding{55}} &  \textcolor{red}{\ding{55}} \\
   Director \cite{hafner2022director} &  \textcolor{red}{\ding{55}} &  \textcolor{red}{\ding{55}} &  \textcolor{green}{\ding{51}} &  \textcolor{red}{\ding{55}} \\
   GC-Director* &  \textcolor{green}{\ding{51}} &  \textcolor{green}{\ding{51}} &  \textcolor{green}{\ding{51}} &  \textcolor{red}{\ding{55}} \\
   Dr. Strategy (\textbf{Ours}) &  \textcolor{green}{\ding{51}} &  \textcolor{green}{\ding{51}} &  \textcolor{green}{\ding{51}} &  \textcolor{green}{\ding{51}} \\

   \bottomrule
   \end{tabular}}
  \caption{A high-level comparison between Dr. ~Strategy and other baselines. *GC-Director is a method we developed based on  \href{https://github.com/danijar/director/tree/main}{the official source code of Director}}
 \label{tab:methods_diff}

\end{table}

\newpage
\section{Additional Experiments}
\arxcg{
\subsection{Sample efficiency comparison between Dr.~Strategy and other baselines}
Given the same sampling budget (number of environment samples) for all baselines, Tables \ref{fig:miniworld_success_table} and \ref{fig:mzkitchen_success_table} show that Dr.~Strategy obtains higher final success rates across most environments compared to other baselines. Moreover, Dr. Strategy shows a faster increment in the performance as shown in Figure \ref{fig:main_results_success_rate}. indicating greater sample efficiency relative to LEXA.

Additionally, Figure \ref{fig:bar_plot_samples} shows the success rates (y-axis) of Dr.~Strategy and other baselines given various sampling budgets (x-axis), highlighting that Dr. Strategy consistently reaches higher success rates in most environments.

\begin{figure}[H]
    \centering
    \includegraphics[width=.9\linewidth]{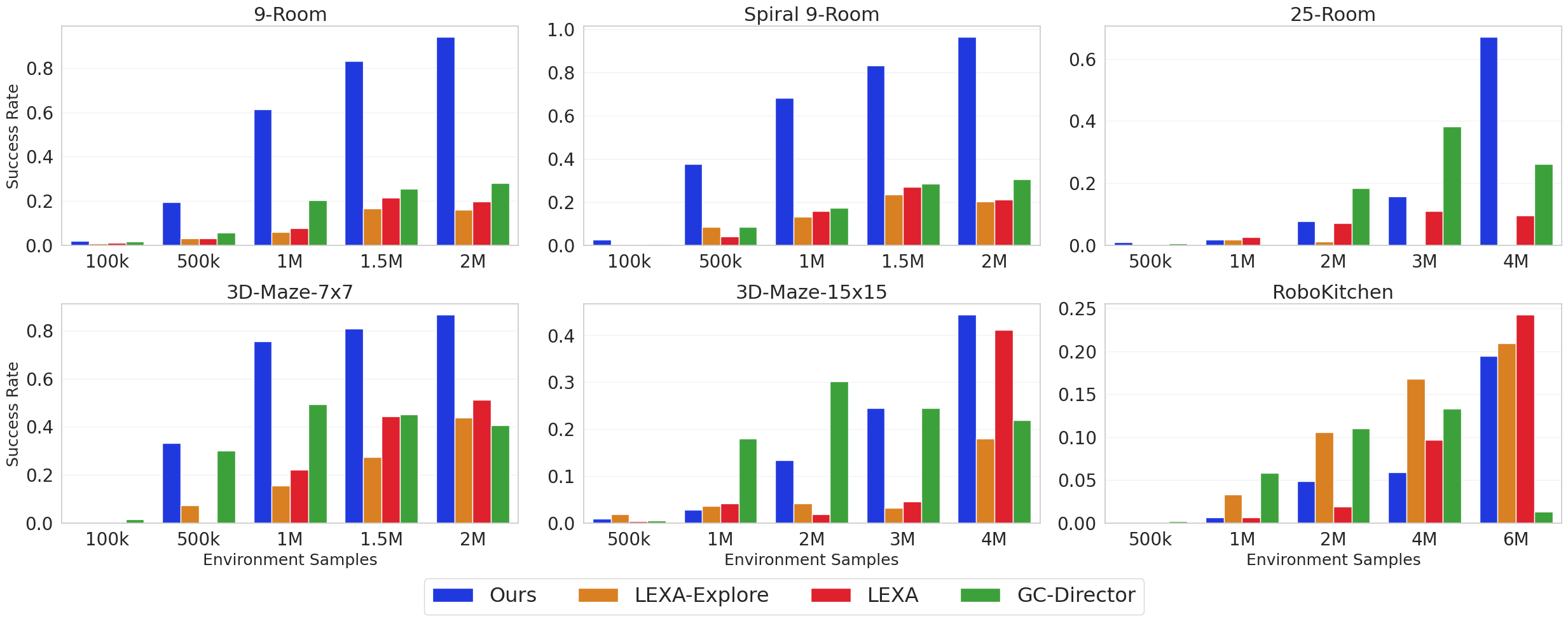}
    \vspace{-1em}
    \caption{\textbf{Success rate given various sampling budgets}. It displays the success rate (y-axis) across various sampling budgets for the baselines} 
    \label{fig:bar_plot_success_rate}
\end{figure}

Figure \ref{fig:bar_plot_samples} further reveals that Dr.~Strategy requires fewer samples to achieve the success rate (x-axis). Furthermore, Dr.~Strategy manages to achieve higher success rates, showcasing its superior performance. In contrast, other baselines fail to achieve the success rate within the training's sampling budget in our experiments.

\begin{figure}[H]
    \centering
    \includegraphics[width=.9\linewidth]{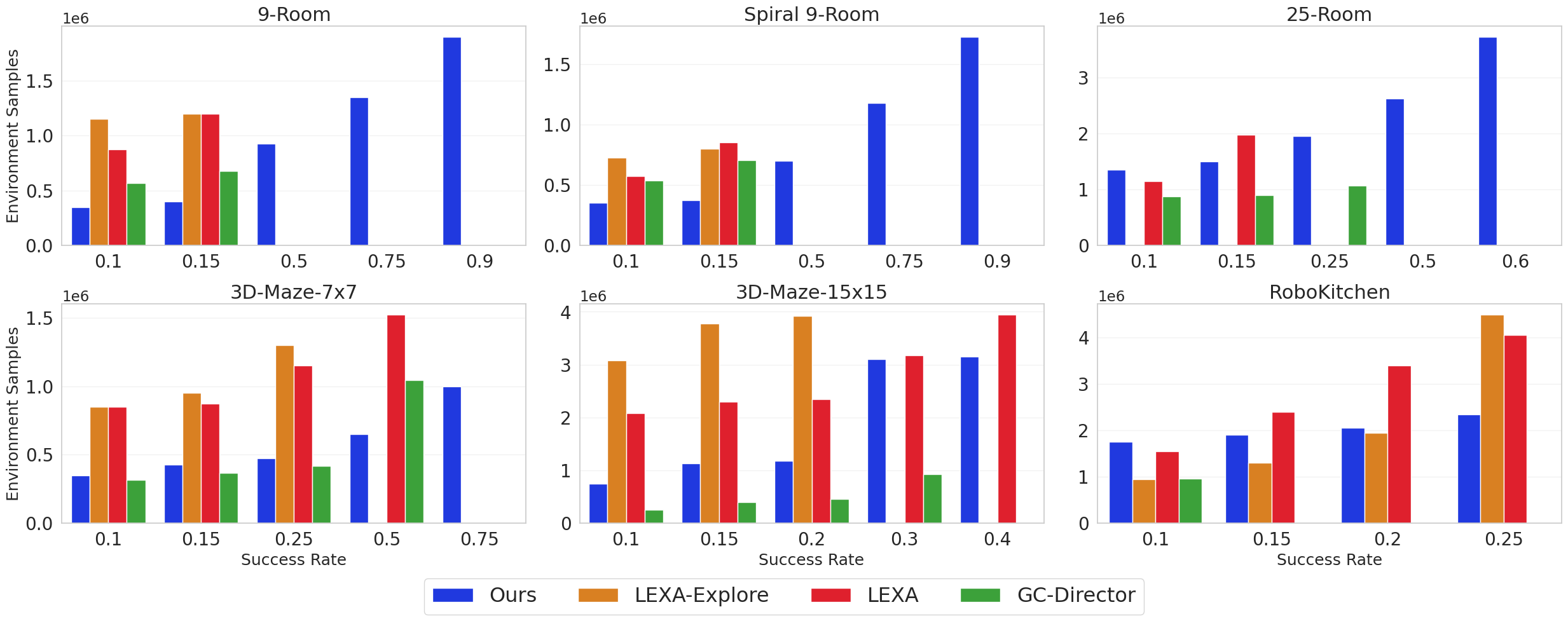}
    \vspace{-1em}
    \caption{\textbf{Number of Samples required to get various success rate thresholds.} It shows the number of environment samples (sampling budget) required to achieve specific success rate thresholds (x-axis). The bar is omitted if the baseline does not achieve the indicated success rate. This omission signifies that the baseline did not achieve the success rate within the given training sampling budget in our experiments} 
    \label{fig:bar_plot_samples}
\end{figure}
}

\subsection{Success rate in RoboYoga Benchmark}
\begin{figure}[H]
    \centering
    \includegraphics[width=.7\linewidth]{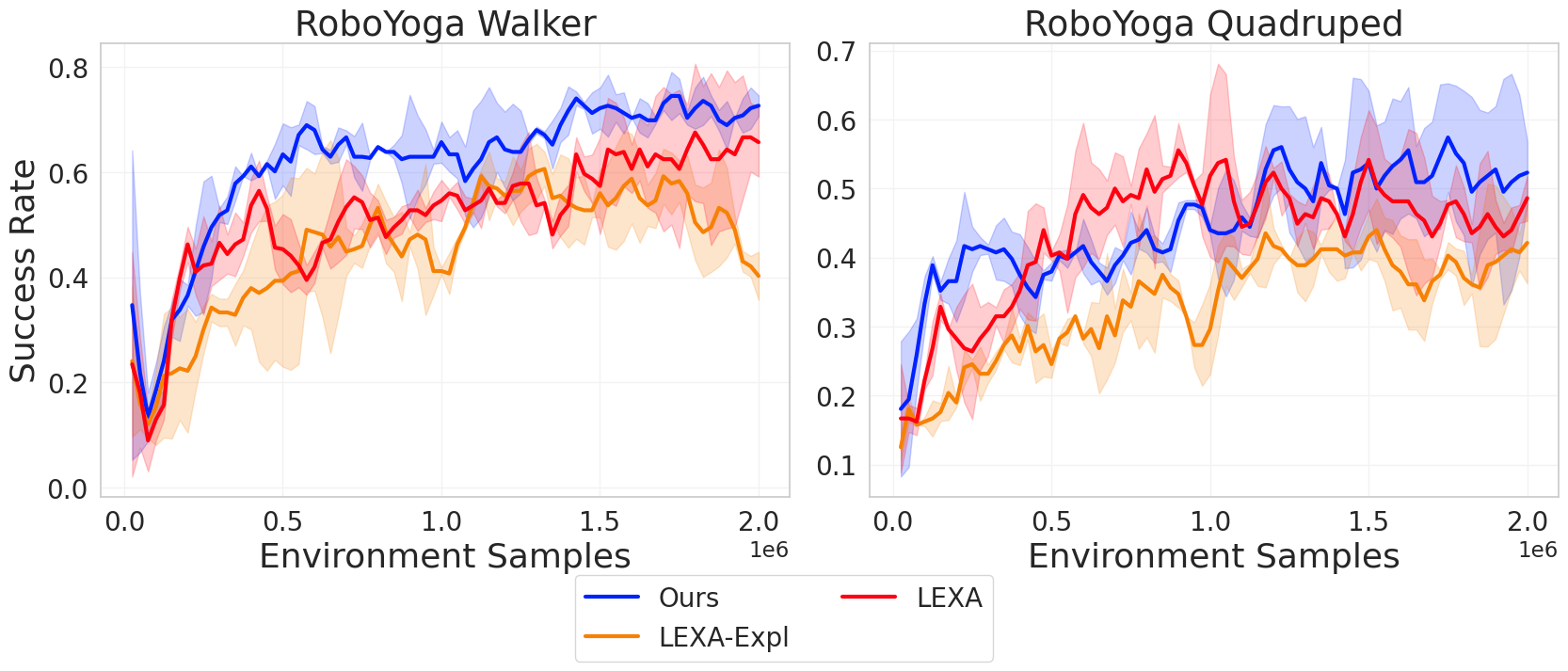}
    \vspace{-1em}
    \caption{\textbf{Zero-shot evaluation of the baselines across RoboYoga Walker and Quadruped}}
    \label{fig:RoboYoga_results}
\end{figure}
 To demonstrate the versatility of our method in various tasks beyond navigation, we evaluate its performance on the RoboYoga benchmark introduced by LEXA \cite{mendonca2021lexa}. To mitigate randomness and noise inherent in the measurements, we adopt the average of three episodes, considering the maximum success achieved in each episode as the performance metric. Specifically, we define the agent's success as achieving the desired goal at least once within an episode. As illustrated in Figure \ref{fig:RoboYoga_results}, our method consistently maintains a commendable level of performance in various domains within the RoboYoga benchmark.

\subsection{Success rate of LEXA-Original and LEXA-Ours}
\label{appendix:LEXA-Original}
\begin{figure}[H]
    \centering
    \includegraphics[width=.7\linewidth]{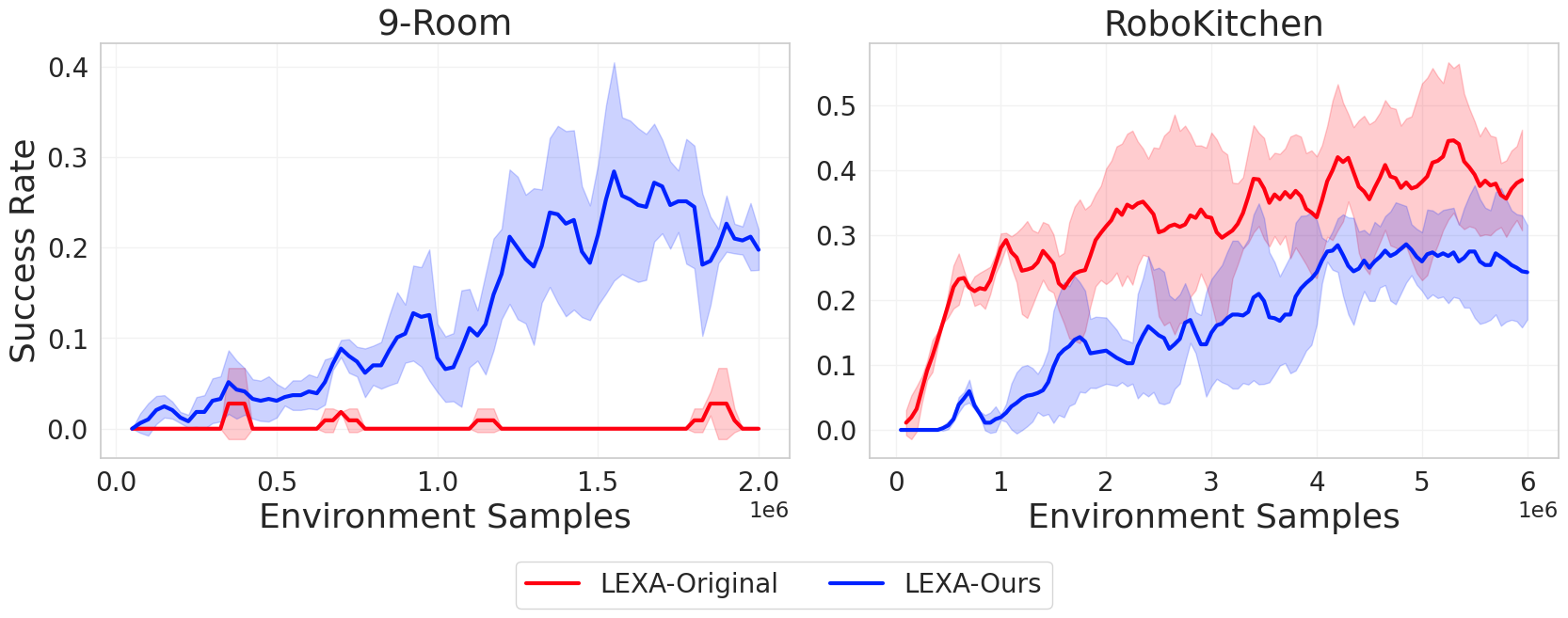}

    \vspace{-1em}
    \caption{\textbf{Success rate of LEXA-Ours and LEXA-Original}} 
    \label{fig:lexa_compare}
\end{figure}
In Figure~\ref{fig:lexa_compare}, we compare our implementation of LEXA (LEXA-Ours) with the original LEXA implementation (LEXA-Original) from \cite{mendonca2021lexa} in 9-room and RoboKitchen. When we run LEXA-Original, we match the configuration and parameters to the original code. For configurations that are not explicit in the original implementation, we match with LEXA-Ours, which is used in Section~\ref{sec:experiments}. The success rate is measured in the same way as mentioned in Appendix~\ref{appendix:env}. Through the success rate, we can clarify that LEXA-Original performance is very low in 9-room. In RoboKitchen, the results of LEXA-Original are similar to the original paper \cite{mendonca2021lexa}. LEXA-Ours show lower performance than LEXA-Original, and the performance gap is around 10\%. This is due to the difference between the implementation mentioned in Appendix~\ref{appendix:baselines}. To compare the model architecture without getting biased by engineering differences, we use LEXA-Ours that uses similar intrinsic reward, world model, and configurations

\newpage
\subsection{Number of Landmarks}
\setlength{\textfloatsep}{10pt plus 1.0pt minus 2.0pt}
\setlength{\intextsep}{10pt plus 1.0pt minus 2.0pt}
\begin{figure}[!htb]
    \centering
    \includegraphics[width=.7\linewidth]{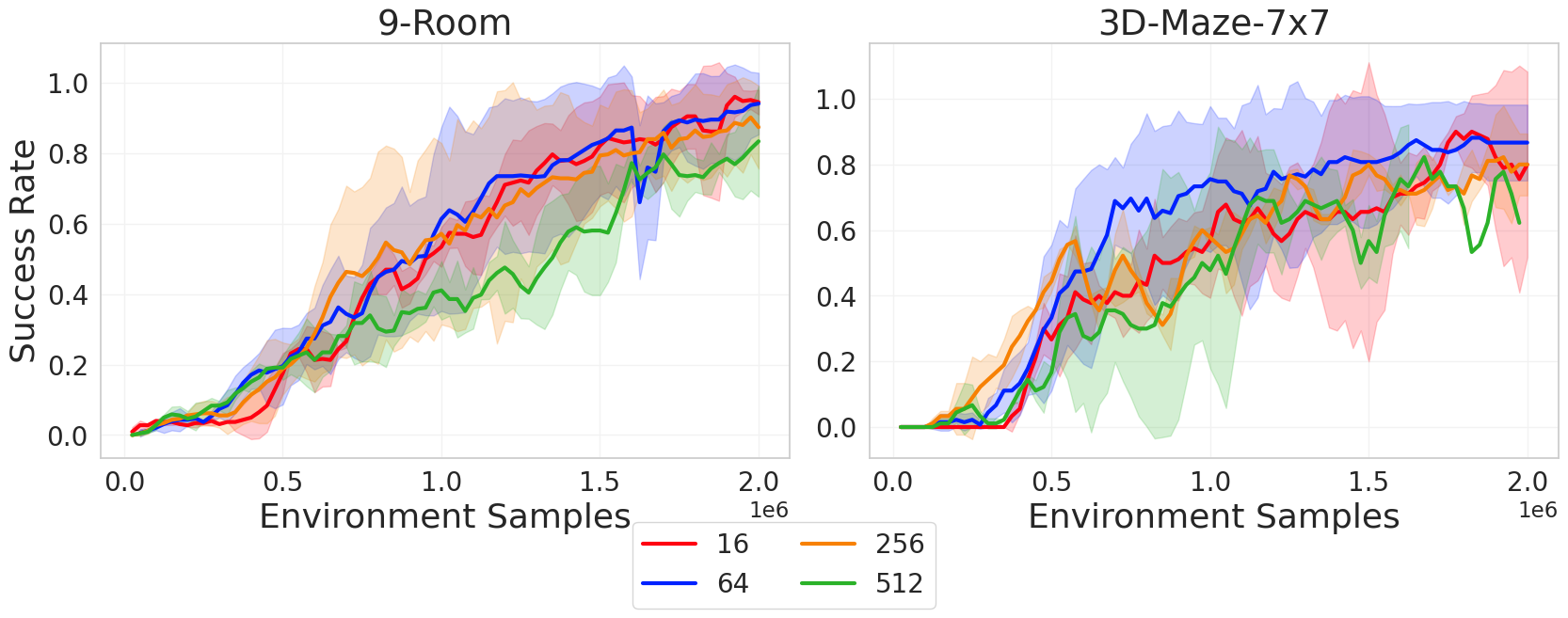}
    \caption{\textbf{Ablation results of using a different number of landmarks (16, 64, 256, 512)}}
    \label{fig:ablation_num_skills}
\end{figure}
 \arxcg{Figure \ref{fig:ablation_num_skills}} shows that increasing the number of landmarks used does not always benefit our method. As 3D-Maze navigation is visually more complex than 2D navigation due to its egocentric observations, it requires a greater number of landmarks to perform the best, which is 64. In 2D navigation (9-rooms) 16 landmarks were enough to have a comparable performance compared to using 64 landmarks. However, using 64 landmarks is able to perform better in some seeds. Using 512 landmarks performs worse than 64 in 9-Room and 3D-Maze-7x7.
 
\subsection{Why is the performance gap of Dr. ~Strategy in Maze-15x15 small compared to Maze-7x7?}
Figure \ref{fig:main_results_success_rate} shows that the performance gap in Maze-15x15 is smaller than that of Maze-7x7. \arxcg{One hypothesis to explain this phenomenon suggests that in Maze-15x15, the larger space and the potential for encountering similar scenes can confuse the agent's ability to generalize from a given goal image. This confusion may arise because larger regions are identified by the same colors. Conversely, Maze-7x7 is smaller, and fewer regions are marked with the same color, as illustrated in Figure \ref{fig:appendix:envs_overview}.}

\begin{figure}[H]
\centering 
    \includegraphics[width=.5\linewidth]{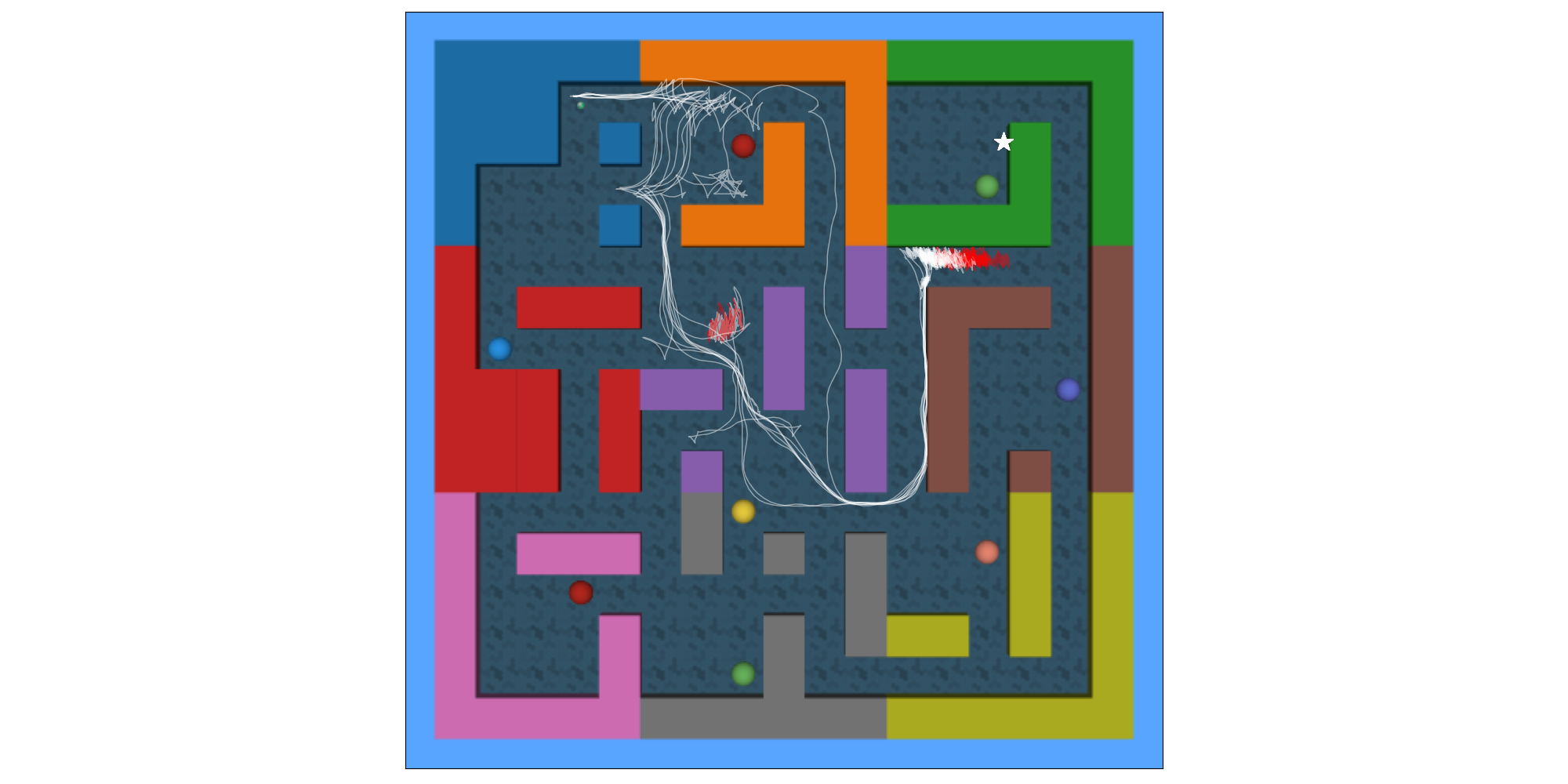}
    \vspace{-1em}
    \caption{\textbf{Visualization of 10 trajectories in 3D-Maze-15x15 of Dr. ~Strategy from the initial state given the green goal in the upper right part.} The trajectories \arxcg{using} highway policy are visualized with white lines, while the trajectories \arxcg{using} achiever are shown with red lines.}
    \label{fig:appendix:mmz15_analysis}
\end{figure}
\vspace{-1em}
\begin{figure}[H]
\centering 
    \includegraphics[width=.7\linewidth]{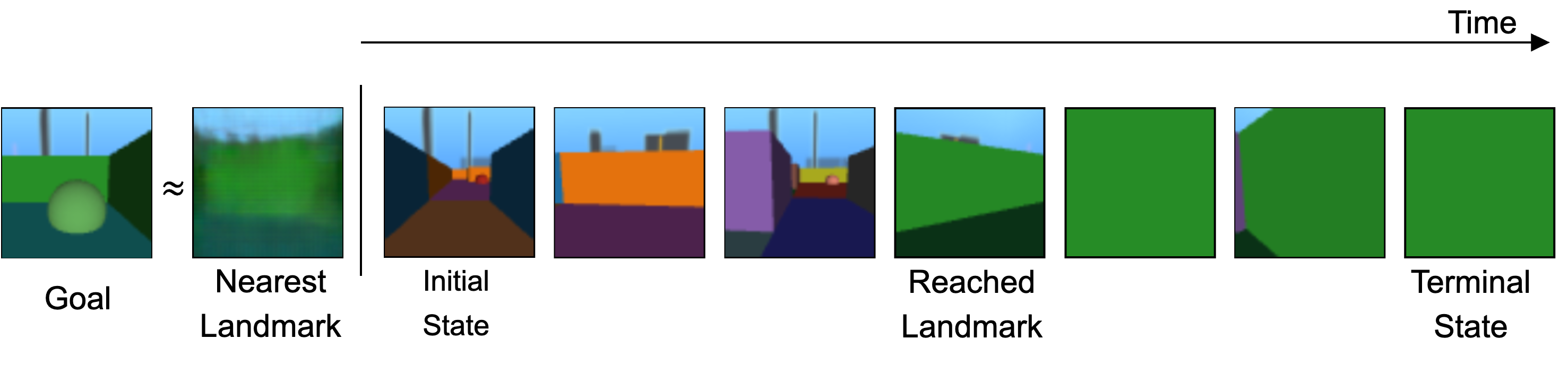}
    \vspace{-1em}
    \caption{\textbf{Visualization of one of the trajectories in Figure~\ref{fig:appendix:mmz15_analysis}.}}
    \label{fig:appendix:mmz15_obs_analysis}
\vspace{-1em}
\end{figure}

  We found that the highway policy given a landmark may sometimes reach a state visually similar to the landmark, but temporally far. As \arxcg{an} empirical evidence, Figure \ref{fig:appendix:mmz15_analysis} shows a top-down view of 10 trajectories for Dr.~Strategy to reach the target in the green room in the upper right part. The agent finds the landmark positioned near the goal denoted by a white star. \arxcg{However, the highway policy could not reach the corresponding landmark within $T_L$ steps, instead the agent stuck at a green wall that is visually similar to the reconstruction of the landmark.} As a result, the agent could not reach the goal which contributes to the agent's low success rate. Figure \ref{fig:appendix:mmz15_obs_analysis} shows the first-person view observation \arxcg{of a trajectory} of the agent.

\subsection{Visualization of Landmarks}
\textbf{2D Navigation}
\begin{center}
\begin{figure}[ht]
    \includegraphics[width=0.49\textwidth, height=0.3\textwidth]{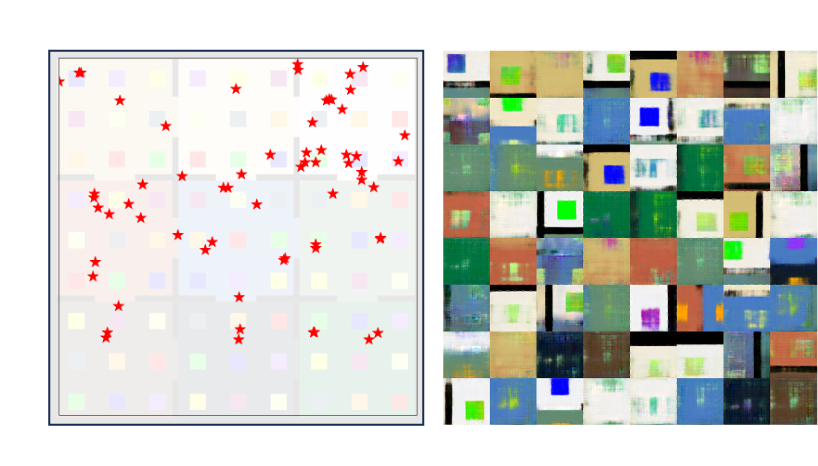}
    \includegraphics[width=0.49\textwidth, height=0.3\textwidth]{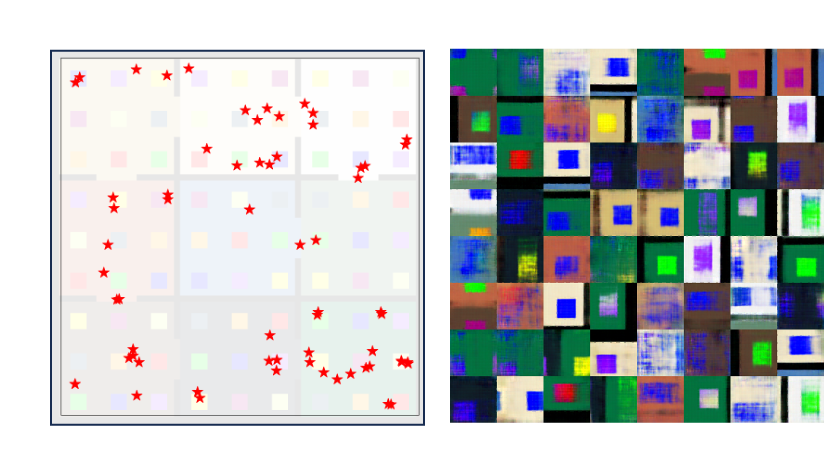}
    \captionsetup{skip=5pt}
    \caption{\textbf{Left:} Landmarks visualizations in 9-room, \textbf{Right:} Landmarks visualizations in Spiral 9-room.}
    \label{fig:2d_navigation_skills}
\end{figure}

\begin{figure}[ht]
    \hspace{-1em}
    \includegraphics[width=\linewidth]{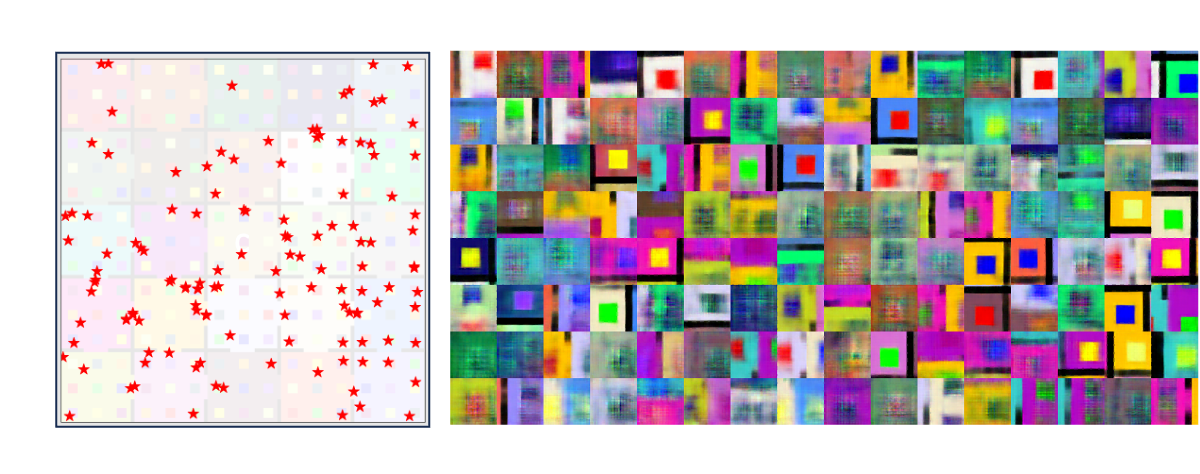}
    \vspace{-1em}
    \caption{Landmarks visualizations in 25-room}
    \label{fig:25room_landmarks_viz}
\end{figure}    
\end{center}

\textbf{3D-Maze}
\begin{center}
\begin{figure}[H]
    \includegraphics[width=0.4\textwidth, height=0.2\textwidth]{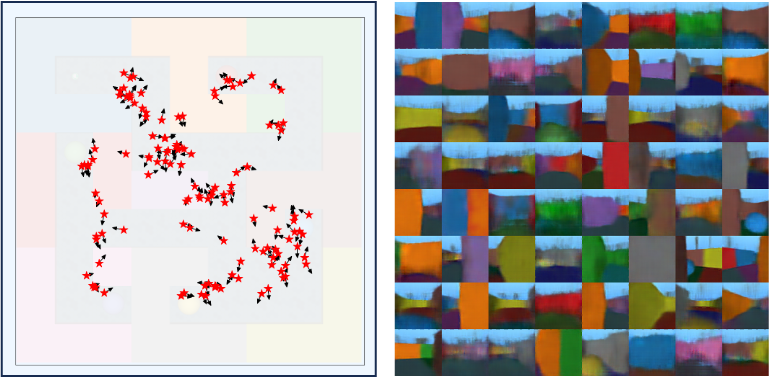}
    \includegraphics[width=0.58\textwidth, height=0.2\textwidth]{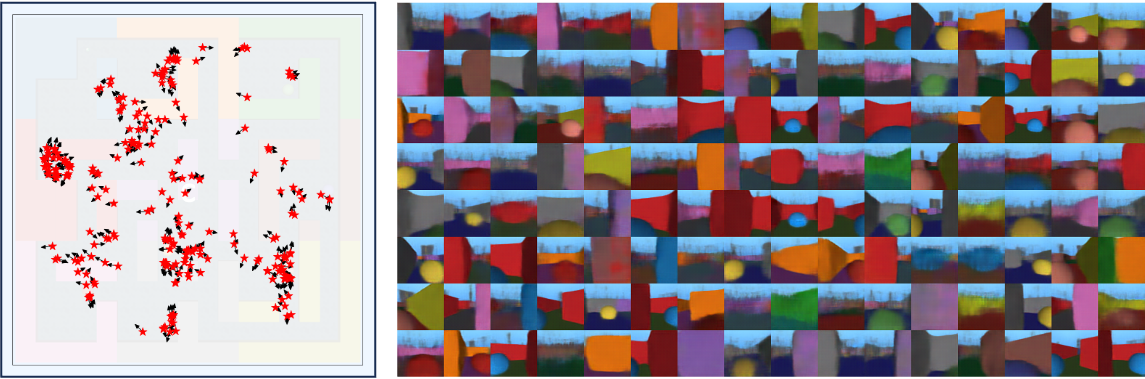}
    \captionsetup{skip=5pt}
    \caption{\textbf{Left:} Landmarks visualizations in 3D-Maze-7x7, \textbf{Right:} Landmarks visualizations in 3D-Maze-15x15.}
    \label{fig:3d_navigation_skills}
\end{figure}
\end{center}

\textbf{RoboKitchen}
\vspace{1em}
\begin{figure}[h]
    \centering
    \includegraphics[width=.7\linewidth]{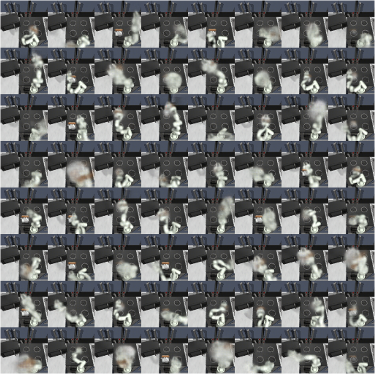}
    \vspace{-1em}
    \caption{Landmarks visualization in RoboKitchen}
    \label{fig:kitchen_landmarks_viz}
\end{figure}
\newpage
\section{Implementation details}
\label{appendix:implementation}

\begin{algorithm}[!ht]
   \caption{Dr. ~Strategy}
    {\bfseries Initialize:} World Model $\mathcal{M}$, Replay buffer $\mathcal{D}$, landmark auto-encoder ($\texttt{enc}_\phi(s), \{l_1, ...., l_N\}, \texttt{dec}_\phi(l)$), Highway policy $\pi_{l}(a_t|s_t, l)$, Explorer $\pi_{e}(a_t|s_t)$, Achiever $\pi_{g}(a_t|s_t, g)$

    \DontPrintSemicolon
        \setlength{\baselineskip}{1.5\baselineskip}
        \tcp{Strategy to Explore}

        \setlength{\baselineskip}{\normalbaselineskip}
        
        \While{\textnormal{exploring}}{
        
        Train $\mathcal{M}$ and landmark auto-encoder on $\mathcal{D}$\;
        
        Train $\pi_{l}$ in imagination of $\mathcal{M}$ to maximize $r_{l}(s_t, l)$ for landmarks $l$ sampled from uniform distribution

        Train $\pi_{e}$ in imagination of $\mathcal{M}$ to maximize exploration reward $r_{e}(s_t)$

        Train $\pi_{g}$ in imagination of $\mathcal{M}$ to maximize $r_{g}(s_t, g)$ for $g = s_{t+H}$
        
        
        
        \setlength{\baselineskip}{1.5\baselineskip}
        \tcp{Curious Landmark}

        \setlength{\baselineskip}{\normalbaselineskip}
        \If{t \textnormal{mod} Pick Curious Landmark every $T_{F}$ steps \textnormal{= 0}}{
            Imagine landmark trajectories $\tau_i$, sampling actions from $\pi_{e}$ starting from $s_{l_i} \sim \texttt{dec}_\phi(l_i), i \in 1,\hdots N $ in parallel
            
            Compute landmark curiosity $C_{i}$ based on $r_{e}$ for each landmark trajectories $\tau_i$
            
            Choose Curious Landmark $l_{C}$ with $p \propto C_{i}$
            
            \setlength{\baselineskip}{1.5\baselineskip}
            \tcp{Landmark-guided Exploration}
    
            \setlength{\baselineskip}{\normalbaselineskip}
            Deploy $\pi_{l}(a_t|s_t,l_C)$ in the environment for $T_L$ steps or until $\lVert s_t - s_{l_C} \rVert < \epsilon$ and grow $\mathcal{D}$, where $s_{l_C} \sim \texttt{dec}_\phi(l_C)$
        }
        Deploy $\pi_{e}(a_t\vert s_t)$ in the environment to explore and grow $\mathcal{D}$
        }
        \setlength{\baselineskip}{1.5\baselineskip}

         \tcp{Strategy to Achieve}

        \setlength{\baselineskip}{\normalbaselineskip}
        \While{\textnormal{evaluating}}{
        \textbf{Given:} Evaluation goal g
        
        \setlength{\baselineskip}{1.5\baselineskip}
        
         \tcp{Find Landmark nearest to goal}

        \setlength{\baselineskip}{\normalbaselineskip}
        
        Imagine trajectory $\tau_g = \{s_0,\hdots s_g\}$, using zero actions, starting from g

        Find Landmark $l_{G}$ nearest to goal where $G = \argmin_{j}\Vert \texttt{enc}_{\phi}(s_g) - l_j \Vert_2$
        
        \setlength{\baselineskip}{1.5\baselineskip}
        \tcp{Focused Achiever}

        \setlength{\baselineskip}{\normalbaselineskip}
        Deploy $\pi_{l}(a_t|s_t,l_{G})$ in the environment for $T_\text{L}$ steps or until $\lVert s_t - s_{l_{G}} \rVert < \epsilon$, where $s_{l_{G}} \sim \texttt{dec}_\phi(l_{G})$ 
        
        Deploy $\pi_{g}(a_t| s_t, g)$ in the environment to reach g.
        }
\end{algorithm}

\vspace{-.2cm}
\textbf{World Model} \quad Following the same architecture as the world model in DreamerV2 \cite{hafner2020dreamerv2}, we use the hyperparameters as indicated in Table \ref{tab:hyperparameters}. 

\arxcg{\textbf{Latent Landmark learning} \quad The latent landmarks are learned through a VQ-VAE, a type of variational autoencoder (VAE) that utilizes vector quantization to obtain a discrete latent representation \cite{van2017vqvae1, razavi2019generating_vqvae2}. VQ-VAE comprises three components: an encoder, a codebook, and a decoder. The encoder projects the input into a latent representation. The codebook learns a discrete set of latent representations known as codes, quantizes the encoder's output by finding the closest code to that output. The decoder then uses the quantized representation to reconstruct the input. A well-known problem in VQ-VAE is code collapse \cite{kaiser2018fast}. To prevent this, we employ the code resampling method mentioned in \cite{mazzaglia2022choreographer}.}

\textbf{Highway policy} \quad is trained in imagination to reach the given landmark. The highway policy is conditioned on the current state and the one-hot representation of the index of the selected landmark from the codebook.

\textbf{Achiever} \quad To train the achiever policy, we use the temporal distance as a reward from \cite{mendonca2021lexa}, the temporal distance prediction network ($\texttt{tdp}$) works in the image embedding space $r_g(e_t, e_g) = -\texttt{tdp}(e_t, e_g)$ and \arxcg{i}ts training is done similarly as in \cite{mendonca2021lexa}. As imagination is in the state space of the world model, we need to decode from the state space to the embedding space, thus we train an embedding decoder network $\texttt{emb}(\hat e_t \mid s_t)$ \arxcg{to predict the image embedding $\hat{e}_t \approx e_t$ given a state $s_t$.}

\textbf{Evaluation and System setup} \quad For the evaluations, we trained all baselines for 3 seeds per environment. The training of our agent took 2 to 6 days based on the environment using 24GB VRAM GPU.

\newpage
\section{Hyperparameters}
\arxcg{Like most other model-based RL methods based on RSSM world models, most of the parameters are set by default to the same values as Dreamer V2 \cite{hafner2020dreamerv2}. We made minor changes only in a few hyper-parameters such as the learning rates of world model, actor, and critic by following the hyperparameters of Choreographer \cite{mazzaglia2022choreographer} as it is also utilizing VQ-VAE like our method. A very small number (only three) of hyperparameters are task-specific as specified in Table \ref{tab:hyperparameters}.}

\vspace{1em}
\label{appendix:hyperparameters}

\begin{table}[ht]
\begin{center}
    \begin{tabular}{ p{8cm}p{3cm}p{3cm}l  }
    \toprule
    \textbf{Name} & \centering \textbf{Symbol} & \centering \textbf{Value} & \\
    \midrule
    \textbf{Latent Landmark} &&& \\
    \midrule
    Number of latent landmarks & \centering $N$ & \centering 64, 128 &\\
    Dimension of latent landmarks & \centering - & \centering 16 &\\
    Commitment loss coefficient & \centering $\beta$ & \centering $10^{-4}$&\\
    Number of layers of latent landmark auto-encoder & \centering - & \centering 4 &\\
    Number of hidden units & \centering - & \centering 400 &\\
    \midrule
    \textbf{World Model} &&& \\
    \midrule
    Replay buffer size & \centering $\mid \mathcal{D} \mid$ & \centering $10^6$ &\\
    Batch size & \centering $B$ & \centering 50 &\\
    Trajectory length & \centering $T_S$ & \centering 50 &\\
    Discrete latent dimensions & \centering - & \centering 32 &\\
    Discrete latent classes & \centering - & \centering 32 &\\
    Number of hidden unit & \centering - & \centering 200 &\\
    KL loss scale & \centering - & \centering 1 &\\
    KL balancing & \centering - & \centering 0.8 &\\
    Learning rate & \centering - & \centering $3 \cdot 10^{-4}$ &\\
    \midrule
    \textbf{Behavior} &&& \\
    \midrule
    Imagination Horizon & \centering $H$ & \centering 15 &\\
    Discount & \centering - & \centering 0.99 &\\
    $\lambda$-target parameter & \centering - & \centering 0.95 &\\
    Actor learning rate & \centering - & \centering $8 \cdot 10^{-5}$ &\\
    Critic learning rate & \centering - & \centering $8 \cdot 10^{-5}$ &\\
    Slow critic update interval & \centering - & \centering 100 &\\
    \midrule
    \textbf{Common} &&& \\
    \midrule
    MLP number of layers & \centering - & \centering 4 &\\
    MLP number of units & \centering - & \centering 400 &\\
    Gradient clipping & \centering - & \centering 100 &\\
    Adam epsilon & \centering - & \centering $10^{-5}$&\\
    Weight decay & \centering - & \centering $10^{-6}$ &\\
    \midrule
    \textbf{Strategy to Explore} &&& \\
    \midrule
    Max. num. of steps to reach Landmark & \centering $T_L$ & \centering 25, 100, 200 & \\ 
    Number of steps to pick the curious landmark & \centering $T_F$ & \centering 150, 500, 1000 & \\
     Reaching landmark threshold & \centering $\epsilon$ & \centering 0.07 & \\

    \bottomrule
    \end{tabular}
    \caption{We use 64 latent landmarks in smaller environments such as 9-room, Spiral 9-room, 3D-Maze 7x7, and Robokitchen. We utilize 128 latent landmarks for larger environments like 25-room and 3D-Maze 15x15. Regarding the exploration strategy, we tailor hyperparameters to each environment. \arxcg{Specifically, for 2D navigation, we set the maximum steps for landmark reaching as 100 and the number of steps to pick the curious landmark at 1000. For 3D-Maze 15x15, these values are adjusted to 200 for landmark reaching and 1000 for curious landmark picking. In the case of 3D-Maze 7x7, we use 100 for landmark reaching and 500 for curious landmark picking. For Robokitchen, the hyperparameters are set at 25 for landmark reaching and 150 for curious landmark picking.}}
    \label{tab:hyperparameters}
    \vspace{-2cm}
\end{center}
\end{table}



\end{document}